\documentclass[10pt,journal,compsoc]{IEEEtran}
\usepackage{multirow}
\usepackage[table]{xcolor}
\usepackage{algorithm} 
\usepackage{amssymb}
\usepackage{amsfonts}
\usepackage{multicol}
\usepackage{graphicx}
\usepackage{cite} 
\usepackage{subfig}
\usepackage{booktabs}
\usepackage[T1]{fontenc}
\usepackage[breaklinks=true,bookmarks=true,pagebackref=false,colorlinks,linkcolor=blue,anchorcolor=red,citecolor=blue,urlcolor=blue]{hyperref}
\usepackage[numbers,sort&compress]{natbib}

\usepackage{comment}
\usepackage{subcaption}
\usepackage{amsmath}
\usepackage{algorithmic}
\usepackage{array}
\usepackage[switch]{lineno}
\usepackage{longtable}
\usepackage{xspace}
\usepackage{url}
\usepackage{overpic}
\usepackage{ragged2e}
\usepackage{framed}
\usepackage{enumitem}

\usepackage{gensymb}
\usepackage{amsmath}
\usepackage{amssymb}
\usepackage{dsfont}
\usepackage{mathrsfs}
\usepackage{yhmath}
\usepackage{MnSymbol}

\usepackage{pifont}
\newcommand{\cmark}{\ding{51}}%
\makeatletter
\DeclareRobustCommand\onedot{\futurelet\@let@token\@onedot}
\def\@onedot{\ifx\@let@token.\else.\null\fi\xspace}

\makeatother

\definecolor{darkgreen}{rgb}{0,0.7,0}
\definecolor{darkblue}{RGB}{31,119,180}
\definecolor{darkred}{RGB}{214,39,40}
\definecolor{mediumgray}{rgb}{0.5,0.5,0.5}
\definecolor{mediumteal}{rgb}{0,0.5,0.5}

\definecolor{ellisred}{rgb}{0.87,0.44,0.38} %
\definecolor{ellisgreen}{rgb}{0.69,0.90,0.52} %
\definecolor{elliscyan}{rgb}{0.29,0.77,0.74} %
\definecolor{ellisorange}{rgb}{0.89,0.55,0.28} %
\definecolor{ellisblue}{rgb}{0.41,0.61,0.86} %

\begin{document}
\title{
Advances in Radiance Field for Dynamic Scene: From Neural Field to Gaussian Field
}

\author{
    Jinlong Fan,
    Xuepu Zeng,
    Jing Zhang,
    Mingming Gong,
    Yuxiang Yang,
    Dacheng Tao

\IEEEcompsocitemizethanks{
\IEEEcompsocthanksitem J. Fan, X. Zeng, and Y. Yang are with the School of Electronics and Information, Hangzhou Dianzi University, Hangzhou, China (e-mail: \{jfan, yyx\}@hdu.edu.cn, 2307665474zxp@gmail.com). %
J. Zhang is with the School of Computer Science, Wuhan University, Wuhan, China (e-mail: jingzhang.cv@gmail.com). %
M. Gong is with Melbourne Centre for Data Science, School of Mathematics and Statistics, University of Melbourne, Parkville, VIC 3010, Australia (e-mail: Mingming.gong@unimelb.edu.au). %
D. Tao is with the College of Computing \& Data Science at Nanyang Technological University, Nanyang Avenue, 639798, Singapore (e-mail: dacheng.tao@gmail.com). 

}%
}

\IEEEtitleabstractindextext{%
\begin{abstract}
\justifying Dynamic scene representation and reconstruction have undergone transformative advances in recent years, catalyzed by breakthroughs in neural radiance fields and 3D Gaussian splatting techniques. While initially developed for static environments, these methodologies have rapidly evolved to address the complexities inherent in 4D dynamic scenes through an expansive body of research. Coupled with innovations in differentiable volumetric rendering, these approaches have significantly enhanced the quality of motion representation and dynamic scene reconstruction, thereby garnering substantial attention from the computer vision and graphics communities. This survey presents a systematic analysis of over 200 papers focused on dynamic scene representation using radiance field, spanning the spectrum from implicit neural representations to explicit Gaussian primitives. We categorize and evaluate these works through multiple critical lenses: motion representation paradigms, reconstruction techniques for varied scene dynamics, auxiliary information integration strategies, and regularization approaches that ensure temporal consistency and physical plausibility. We organize diverse methodological approaches under a unified representational framework, concluding with a critical examination of persistent challenges and promising research directions. By providing this comprehensive overview, we aim to establish a definitive reference for researchers entering this rapidly evolving field while offering experienced practitioners a systematic understanding of both conceptual principles and practical frontiers in dynamic scene reconstruction.
We maintain an active repository of literature and open-source implementations to complement this survey at 
%
\href{https://github.com/MiliLab/Awesome-DynRF}{Awesome-DynRF}.

\end{abstract}

}

\maketitle
\IEEEdisplaynontitleabstractindextext
\IEEEpeerreviewmaketitle

\section{Introduction}
\label{sec:introduction}

\begin{figure*}[t]
    \centering
    \begin{overpic}[width=0.95\linewidth]{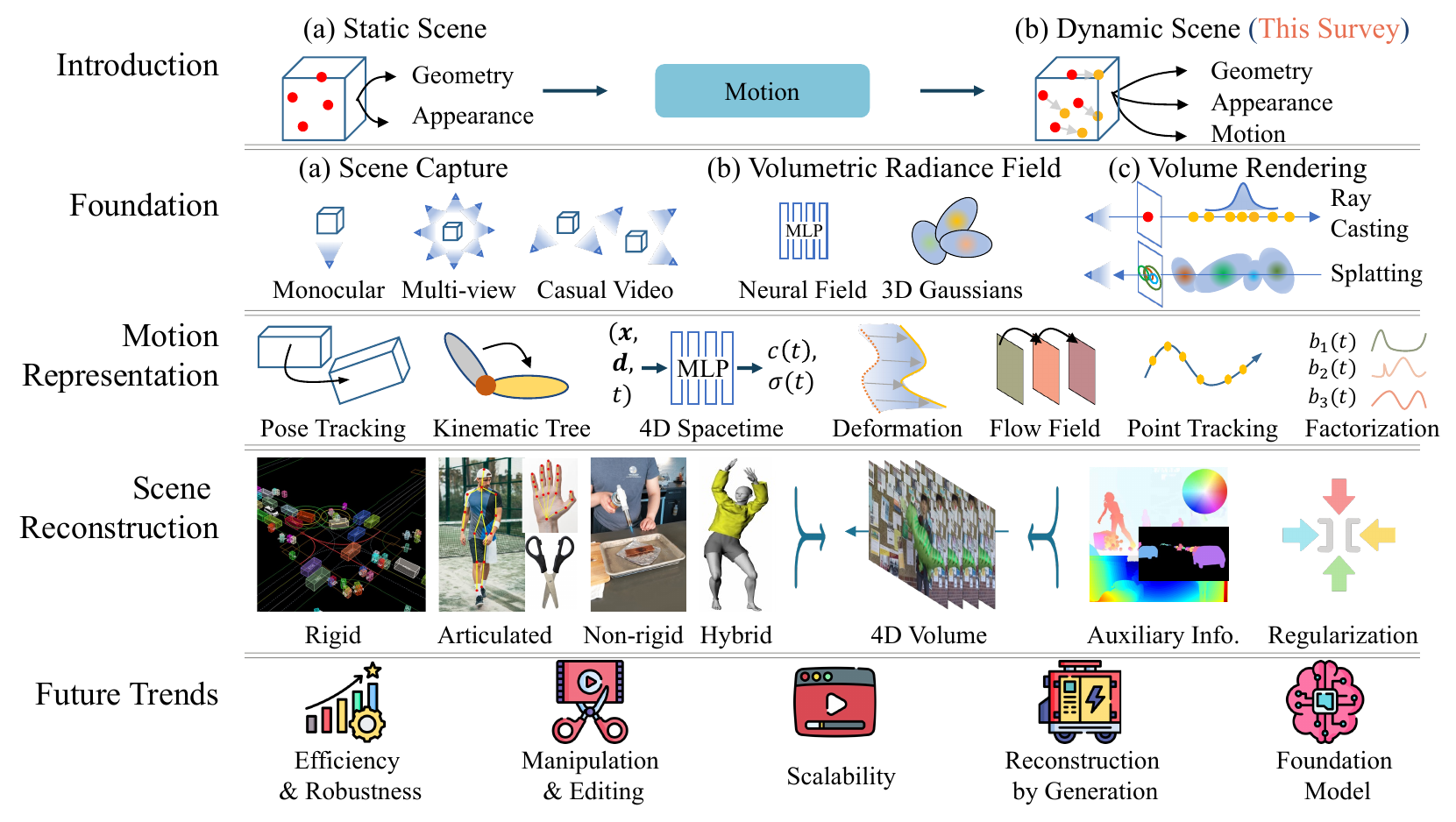}
    \put(7,49){Sec.~\ref{sec:introduction}}
    \put(7,39){Sec.~\ref{sec:foundational_concepts}}
    \put(7,28){Sec.~\ref{sec:motion_repre}}
    \put(7,17){Sec.~\ref{sec:scene_repre}}
    \put(7,6){Sec.~\ref{sec:future}}
    \end{overpic}
    
    \caption{\textbf{Survey at A Glance.} 
    \textbf{(a) Introduction and Foundation.} We trace the evolution from static to dynamic scene representation, highlighting the challenges of jointly modeling motion, geometry, and appearance using radiance fields.
    \textbf{(b) Motion Representation.} We categorize motion patterns and their representation paradigms, examining how they enable complex motion modeling while addressing inherent limitations.
    \textbf{(c) Scene Reconstruction.} We analyze how motion representations enable scene reconstruction, discussing these methods within a unified framework while investigating how auxiliary information and regularization strategies constrain the learning of radiance fields.
    \textbf{(d) Future Trends.} We explore promising research directions and how dynamic scene reconstruction could benefit by aid of the rapid development
    of foundation models and large language models.
    }
    \label{fig:overview}
\end{figure*}
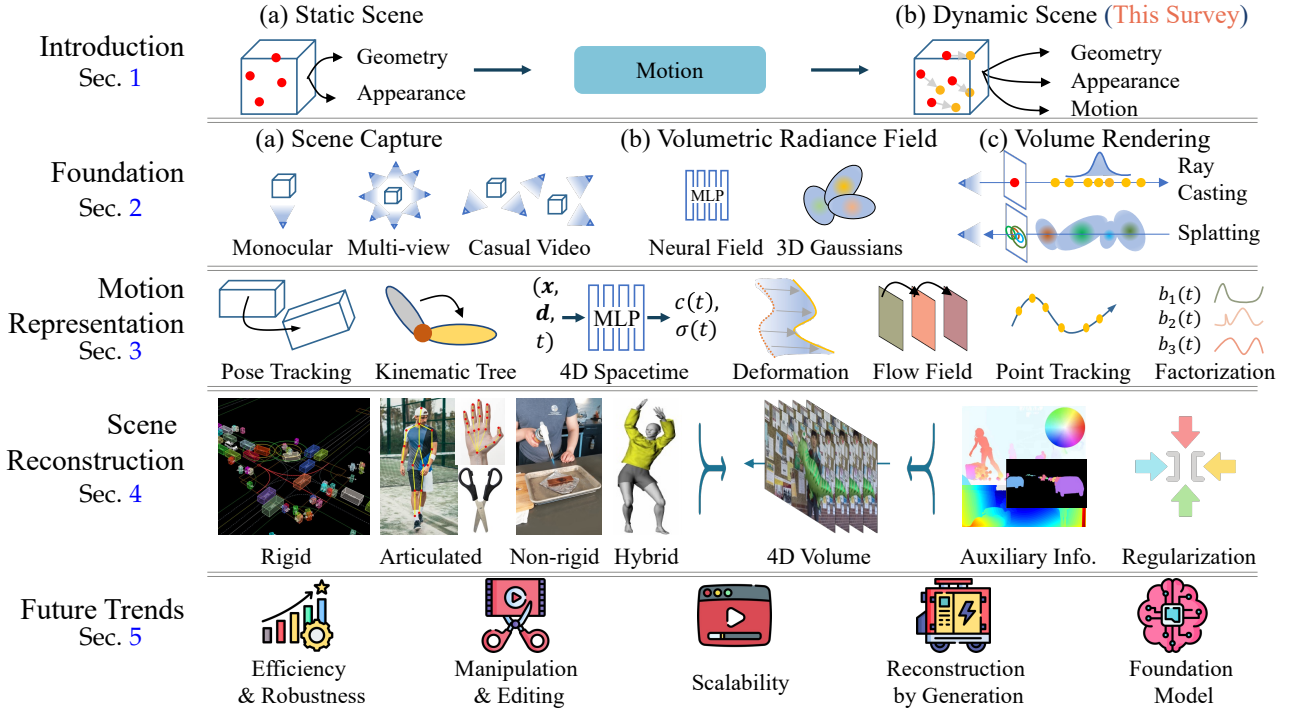

\IEEEPARstart{S}{cene} representation constitutes a fundamental cornerstone in computer vision, with robust 3D scene reconstruction remaining an enduring and vibrant research domain for decades. Recent computational paradigms, catalyzed by advances in 3D representation and differentiable rendering, have reinvigorated methodologies that capture and reconstruct the intricate details of real-world environments. Among these developments, radiance fields have emerged as pivotal representations in 3D vision, particularly through milestone approaches proposed in Neural Radiance Fields (NeRF)~\cite{mildenhall2021nerf} and 3D Gaussian Splatting (3DGS)~\cite{kerbl3Dgaussians}. By coupling these fields with differentiable volumetric rendering~\cite{niemeyer2020differentiable}, analysis-by-synthesis methods have achieved unprecedented fidelity in static scene reconstruction.

However, these early successes predominantly addressed static settings~\cite{mildenhall2021nerf,kerbl3Dgaussians,muller2022instant,chen2022tensorf,barron2021mip,yu2024mip}, overlooking the inherent dynamics of real-world scenes. In practice, virtually every environment exhibits temporal evolution, whether from object movement, changing illumination, or evolving scene geometry. Recognizing this limitation, numerous recent approaches have extended static radiance field frameworks to handle dynamic scenes and accommodate complex temporal variations~\cite{pumarola2021d,park2021nerfies,park2021hypernerf,wu20244d,yang2024deformable}. This rapidly expanding corpus of techniques underscores the necessity for a comprehensive survey summarizing the state-of-the-art in dynamic scene representation and reconstruction.

The principal challenge in dynamic scene reconstruction lies in accurately modeling the temporal dimension--the motion field. Motion representation constitutes the cornerstone of dynamic scene reconstruction, where the precision of point correspondence across frames directly determines the quality of recovered dynamic content. To address this challenge, our survey begins by systematically examining the taxonomy of motion types and comprehensively reviewing strategies to represent these various motions in 3D space. Recent advances have demonstrated that continuous and flexible formulations prove especially effective in faithfully representing complex motions without relying on oversimplified discretizations, as evidenced by innovations in neural scene flow fields, deformable radiance fields, and 4D neural volumes~\cite{liNeuralSceneFlowFieldsSpacetime2021,park2021nerfies,fridovich2023k,caoHexplaneFastRepresentationDynamicScenes2023,wu20244d,yang2024deformable}.  

Building upon these motion representations, we analyze diverse strategies for reconstructing and rendering dynamic scenes under various motion conditions from multiple input modalities, including monocular video, multi-view video, and casually captured one. We propose examining these methods from a unified perspective, wherein any dynamic scene can be conceptualized as a static reference space coupled with an appropriate motion representation addressing specific motion types. Furthermore, a significant challenge in this domain involves disentangling the inherent ambiguities between motion, geometry, and appearance. To overcome these ambiguities, researchers frequently employ auxiliary information and regularization techniques as additional supervision or constraints, guiding solutions toward physically plausible and realistic dynamic reconstructions~\cite{liu2023robust,luiten2024dynamic,wangTrackingEverythingEverywhereAllOnce2023,wangFlowSupervisionDeformableNerf2023,liNeuralSceneFlowFieldsSpacetime2021,li2023dynibar,fu2022panoptic,kundu2022panoptic,wang2021neural}.  

This survey aims to chart the evolutionary trajectory of dynamic scene representation, highlighting the substantial progress enabled by neural radiance fields and 3D Gaussian splatting while drawing attention to persistent challenges that require further investigation. Fig~\ref{fig:overview} provides a comprehensive overview of this survey's structure and scope. By offering a systematically organized examination of recent innovations in motion representation and dynamic scene reconstruction, we seek to provide both newcomers and experienced researchers with valuable insights into emerging directions where dynamic scene modeling can evolve, ultimately facilitating increasingly realistic, interactive, and robust applications across computer vision, graphics, and related fields.

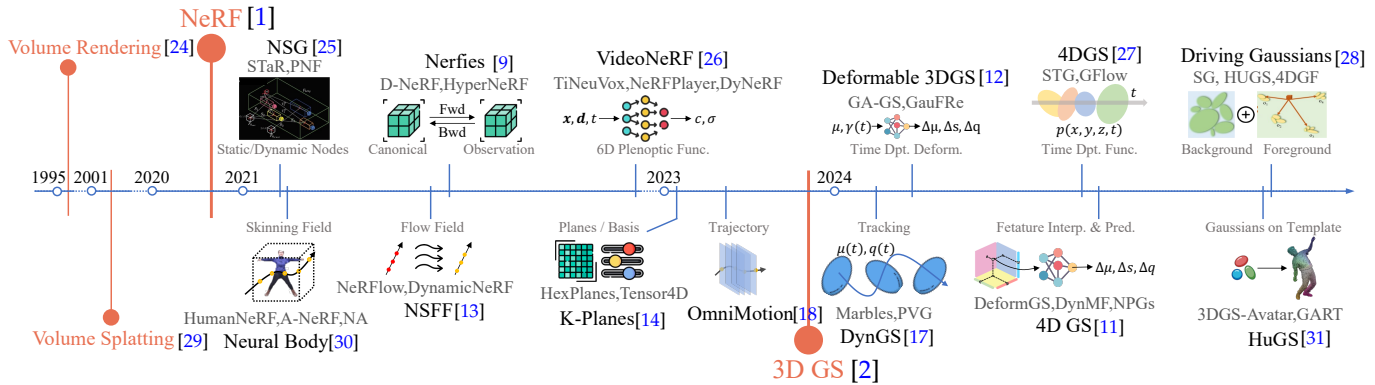
\begin{figure*}[t]
    \centering
    \begin{overpic}[width=1.0\linewidth]{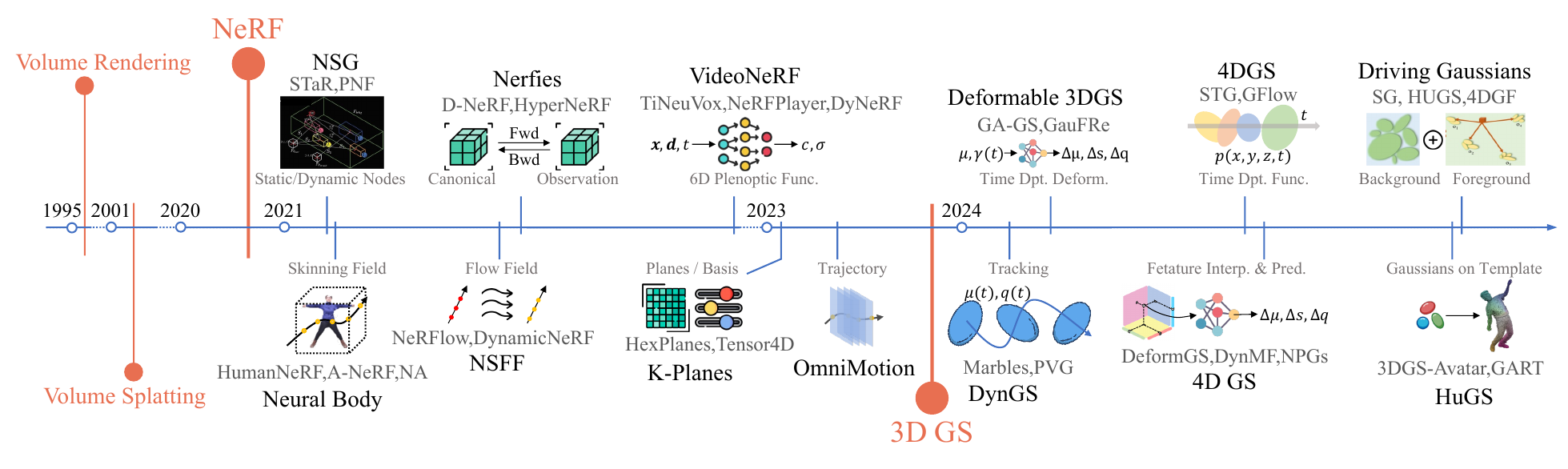}
    
\put(12.0,25.5){\scriptsize~\cite{max1995optical}}
\put(17.9,27.5){~\cite{mildenhall2021nerf}}
\put(22.8,25.5){\scriptsize~\cite{ostNeuralSceneGraphsDynamicScenes2021}}
\put(36,24.3){\scriptsize~\cite{park2021nerfies}}
\put(51,24.6){\scriptsize~\cite{xian2021space}}
\put(71.5,23.3){\scriptsize~\cite{yang2024deformable}}
\put(81.2,25){\scriptsize~\cite{yangreal}}
\put(97.5,24.7){\scriptsize~\cite{zhou2024drivinggaussian}}

\put(13, 4){\scriptsize~\cite{zwicker2001ewa}}
\put(24, 3.9){\scriptsize~\cite{peng2021neural}}
\put(33.2, 6.2){\scriptsize~\cite{liNeuralSceneFlowFieldsSpacetime2021}}
\put(46.4, 5.5){\scriptsize~\cite{fridovich2023k}}
\put(57.9, 6){\scriptsize~\cite{wangTrackingEverythingEverywhereAllOnce2023}}
\put(62, 1.7){~\cite{kerbl3Dgaussians}}
\put(66, 4.3){\scriptsize~\cite{luiten2024dynamic}}
\put(79.9, 5){\scriptsize~\cite{wu20244d}}
\put(95, 4.2){\scriptsize~\cite{kocabas2024hugs}}
\end{overpic}
\caption{\textbf{Roadmap of Dynamic Scenes in Radiance Fields.} 
This chronological timeline illustrates the evolution of the field, organizing works into methodological clusters based on their representation paradigms. The representative or first work within each cluster appears in black with accompanying paradigm illustrations, while the dates of remaining works may vary within clusters. Seminal contributions that significantly advanced the field are highlighted with colors.
}
    \label{fig:roadmap}
\end{figure*}

\subsection{Roadmap}
\label{sec:roadmap}
Fig.~\ref{fig:roadmap} illustrates the chronological evolution of dynamic scene representation in radiance fields. The journey begins with differentiable volume rendering~\cite{max1995optical}, which enabled gradient-based optimization of 3D-to-2D transformations. A breakthrough came with NeRF~\cite{mildenhall2021nerf}, which could learn scene representations from only 2D images without 3D supervision. Early dynamic extensions utilized scene graphs with static and dynamic nodes to handle rigid objects~\cite{ostNeuralSceneGraphsDynamicScenes2021,yuan2021star,kundu2022panoptic}, while parametric template-based approaches (using SMPL~\cite{loper2015smpl} or MANO~\cite{romero2017embodied}) achieved significant success in human avatar reconstruction~\cite{peng2021neural,weng2022humannerf,liu2021neural}. For general dynamic scenes, two main paradigms emerged: methods like Nerfies~\cite{park2021nerfies} that linked observation space to a canonical space via deformation fields~\cite{park2021hypernerf,pumarola2021d}, and approaches using frame-to-frame flow fields to establish temporal connections~\cite{liNeuralSceneFlowFieldsSpacetime2021,gaoDynamicViewSynthesisDynamicMonocular2021a,duNeuralRadianceFlow4dView2021,li2023dynibar}. Subsequently, researchers explored integrated 4D spacetime representations~\cite{xian2021space,fangFastDynamicRadianceFieldsTimeaware2022,songNerfplayerStreamableDynamicSceneRepresentation2023a,li2022neural}, factorization methods for efficient field modeling~\cite{fridovich2023k,caoHexplaneFastRepresentationDynamicScenes2023,shao2023tensor4d}, and techniques like OmniMotion~\cite{wangTrackingEverythingEverywhereAllOnce2023} that leveraged long-term dense point tracking.  

The field evolved further with the advent of 3DGS~\cite{kerbl3Dgaussians}, which represents 3D scene with Gaussian primitives and is rendered through efficient splatting techniques~\cite{zwicker2001ewa}. Replacing implicit neural representations with explicit Gaussian primitives, 3DGS demonstrates great potential for motion modeling. The approaches, that utilizing 3DGS to represent dynamic scenes, fall into several categories: methods that initialize Gaussians at time $t_0$ as a canonical space and warp them using time-dependent deformation fields~\cite{yang2024deformable,liang2025gaufre,lu3dGeometryawareDeformableGaussianSplatting2024}; techniques that track Gaussian primitive movements to represent dense motion fields~\cite{luiten2024dynamic,stearns2024dynamic,chen2023periodic}; approaches employing time-dependent functions to characterize varying Gaussian properties; and methods utilizing interpolated time-related features on factorized feature planes to predict dynamic Gaussian properties (position, scale, and rotation)~\cite{wu20244d,duisterhof2023deformgs}. Recent advances have also applied 3D Gaussian fields to represent human avatars~\cite{kocabas2024hugs,lei2024gart,qian20243dgs} and scenes with rigid objects~\cite{zhou2024drivinggaussian,zhou2024hugs,fischer2024dynamic,yan2024street}, enabling part-level animation or object-level manipulation.

\subsection{Comparison to Related Surveys}
\label{sec:related-survey}

Rapid progress in differentiable rendering and radiance field has led to numerous surveys in this field. The works of \cite{gaoBriefReviewDifferentiableRenderingRecent2024} provide foundational insight into differentiable rendering, while surveys in~\cite{xieneuralfieldsvisualcomputing2022b,gaoNerfNeuralRadianceField3d2022,xieneuralfieldsvisualcomputing2022b} document neural radiance field variants. More recent surveys on 3DGS~\cite{chen2024survey,wuRecentAdvances3dGaussianSplatting2024,bao20253d} have emerged to capture developments in explicit representations.

However, a critical gap persists: existing surveys predominantly address static scene representation, with only peripheral coverage of dynamic scenes. Domain-specific surveys~\cite{tosiHowNerfs3dGaussianSplatting2024,slapakNeuralRadianceFieldsIndustrialRobotics2024,wangNeRFRoboticsSurvey2024,sunHuman3dAvatarModelingImplicit2022,ming2025benchmarking} incorporate dynamic aspects, but focus on application-specific challenges rather than fundamental problems in dynamic scene representation. Even surveys explicitly addressing non-rigid reconstruction either take a broader view beyond neural fields~\cite{yunusRecentTrends3DReconstructionGeneral2024,tretschkstateartdensemonocularnonrigid2023} or focus narrowly on specific scenarios\cite{gu20253d,liu2024survey}. Our survey distinguishes itself by providing a comprehensive analysis specifically dedicated to dynamic scene reconstruction using radiance fields, including NeRF and 3DGS. We uniquely bridge various representation paradigms within a unified framework, offering a new perspective on this rapidly evolving field.

\subsection{Contributions}
\label{sec:contribution}
To summarize, this survey has three key contributions:
(a) We present a structured roadmap tracing dynamic scene representation from NeRF to 3D Gaussian Splatting, establishing a unified taxonomy of approaches organized by motion types and representation paradigms. This integrated perspective reveals critical connections between methodological clusters that isolated technical reviews often overlook.
(b) We identify fundamental challenges in representing diverse motion patterns based on our survey of over 200 papers. Our analysis examines how different motion representation paradigms address these challenges and how auxiliary information and regularization techniques enhance reconstruction quality and temporal consistency.
(c) We analyze how recent breakthroughs in generative and foundation models have transformed the trajectory of dynamic scene reconstruction. To support ongoing research, we maintain an actively updated repository documenting emerging methods, open-source implementations, and benchmark results across the spectrum of dynamic scene representation approaches.

\section{Foundational Concepts and Knowledge}
\label{sec:foundational_concepts}

\subsection{Scene Capture}

\subsubsection{Sensor Types}
Scene capture relies on various sensor types, each with distinct characteristics. \textbf{RGB cameras} are the most accessible and widespread sensors, providing dense color information but lacking direct depth measurements, while \textbf{RGB-D sensors} enhance this capability by combining RGB data with depth information to simplify 3D reconstruction, though they often suffer from limited range, noise, and sensitivity to environmental conditions. \textbf{LiDAR} systems employ laser pulses to generate precise point clouds with accurate geometry, but the resulting data is typically sparse and may require alignment with RGB images before use as auxiliary information in reconstruction pipelines.

\subsubsection{Capture Setting}
We categorize scene capture settings into three distinct classes: \textbf{monocular capture} which encompasses both strict monocular with stationary cameras and effective multi-view when camera motion is comparable to object speed, \textbf{multi-view capture} that employs multiple synchronized cameras that simultaneously observe the scene from different angles, providing comprehensive geometric constraints, and \textbf{casual video capture}, footage obtained from handheld devices in unconstrained environments without professional setups. The key distinction between these approaches lies in their spatiotemporal sampling characteristics and the relative motion between camera and scene objects~\cite{gaoMonocularDynamicViewSynthesisReality2022}. While multi-view setups offer superior reconstruction fidelity through comprehensive spatial coverage with minimal occlusions, monocular methods can achieve reasonable results when relative camera-object motion is appropriately balanced, and casual video approaches trade reconstruction quality for accessibility and flexibility in everyday scenarios.

\subsection{Volumetric Radiance Field}

\subsubsection{Neural Radiance Field}
NeRF~\cite{mildenhall2021nerf} represents a scene as a continuous 5D function parameterized by a Multi-Layer Perceptron (MLP) with parameters $\theta$, formulated as:
 \begin{equation}
    \mathcal{F} _{\theta}(\mathbf{x},\mathbf{d}) \longmapsto (\mathbf{c},\sigma),
\end{equation}
where $\mathbf{x}=(x,y,z)$ denotes spatial coordinates, and $\mathbf{d}$ represents viewing directions as normalized unit vectors from the camera's optical center to pixel positions. The network outputs RGB color $\mathbf{c}=(r,g,b)$ and volume density $\sigma$, where density represents view-independent geometry, while color varies with viewing direction to model view-dependent effects such as specular highlights.

\textbf{Volume Rendering.} The rendering process in NeRF employs ray tracing principles~\cite{max1995optical}, integrating color and density values along camera rays $\mathbf{r}(t)=\mathbf{o}+t\mathbf{d}$ to produce the final pixel color. This integration is expressed in discrete form:
\begin{equation}
    \hat{C}(\mathbf{r}) = \sum_{i=1}^{N} \alpha_i T_i \mathbf{c}_i,
    \label{eq:rendering}
\end{equation}
where $T_i = \exp\left(-\sum_{j=1}^{i-1} \sigma_j \delta_j \right)$, $\delta_i$ denotes the distance between adjacent samples, and $\alpha_i = 1 - \exp(-\sigma_i \delta_i)$ represents the opacity at each sample point.

\subsubsection{3D Gaussian Splatting}
3DGS~\cite{kerbl3Dgaussians} provides an alternative to employs explicit, learnable primitives rather than implicit neural networks to represent the radiance field. This method represents scenes as collections of anisotropic 3D Gaussians $\mathcal{G}$, each parameterized by its position $\mathbf{\mu} \in \mathbb{R}^3$, covariance matrix $\mathbf{\Sigma} \in \mathbb{R}^{3\times 3}$, opacity $o \in [0,1]$, and color attributes $\mathbf{c}$. The covariance matrix, defining the Gaussian's shape and orientation, is constructed from a scaling factor $\mathbf{S}\in \mathbb{R}^{3}$ and rotation matrix $\mathbf{R} \in \mathbb{R}^{3\times 3}$ as $\mathbf{\Sigma}=\mathbf{R}\mathbf{S}\mathbf{S}^\top\mathbf{R}^\top$. The color attributes are typically represented by spherical harmonics (SH) coefficients to model view-dependent appearance effects. All these properties are learnable parameters that are optimized through gradient descent to align with observations.

\textbf{Volume Splatting.} Unlike NeRF's ray marching approach, Gaussian Splatting employs a tile-based rasterization pipeline for efficient rendering~\cite{zwicker2001ewa}. The process involves projecting 3D Gaussian primitives onto the 2D image plane, a technique commonly referred to as "splatting." When projecting each 3D Gaussian, its center is transformed to the 2D image space as $\mathbf{\mu}^{2D} = \mathbf{J}\mathbf{W}\mathbf{\mu}$, and its covariance matrix is similarly transformed to $\mathbf{\Sigma}^{2D}=\mathbf{J}\mathbf{W}\mathbf{\Sigma}\mathbf{W}^\top\mathbf{J}^\top$, where $\mathbf{W}$ represents the viewing transformation matrix that maps from world to camera coordinates, and $\mathbf{J}$ is Jacobian of the affine approximation of projective transformation.

The final color of each pixel is computed by blending all Gaussian splats that overlap at that pixel location. These Gaussians are first sorted by depth to ensure proper occlusion handling, then composited in front-to-back order according to the equation:
\begin{equation}
    \mathbf{c} \;=\; \sum_{i \in \mathcal{N}} \mathbf{c}_i\,\alpha_i^{2D} \prod_{j=1}^{i-1} \bigl(1 - \alpha_j^{2D}\bigr),
    \label{eq:splatting}
\end{equation}
where $\mathcal{N}$ represents the set of Gaussians contributing to the pixel, $\mathbf{c}_i$ is the color of the $i$-th Gaussian, and $\alpha_i^{2D}$ is the 2D projected alpha value of each gaussian primitive.

\section{Dynamic Motion Representation}
\label{sec:motion_repre}

\begin{figure}
    \centering
    \includegraphics[width=1.0\linewidth]{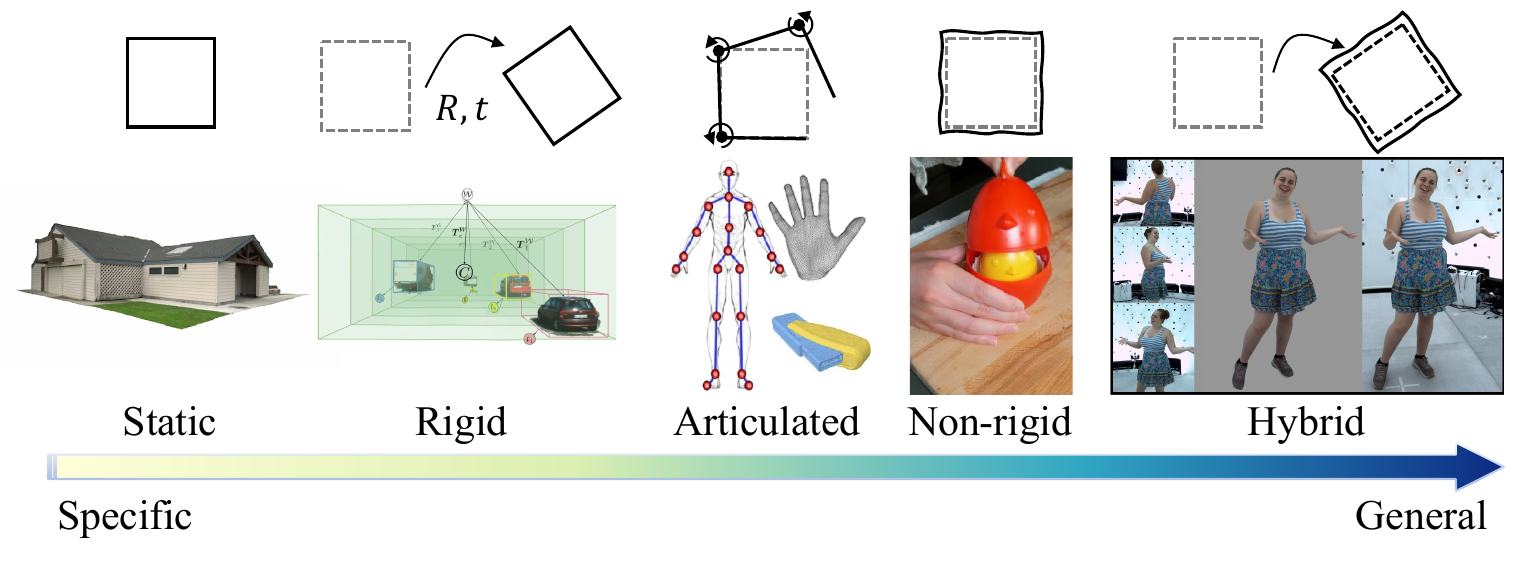}
    \caption{A 2D illustration of various motion types.}
    \label{fig:motion}
\end{figure}

Accurately modeling dynamic motion forms a critical foundation for scene reconstruction, understanding, and analysis. Real-world environments exhibit diverse motion patterns that can be categorized hierarchically from specific to general types. We classify these patterns into rigid motion, articulated motion, general non-rigid motion, and hybrid motion, which combines multiple patterns, as illustrated in Fig.~\ref{fig:motion}. Although articulated motion represents a specific type of non-rigid movement, we address it separately due to its distinct representation approaches \cite{aggarwal1994articulated}. Throughout this survey, we use the term ''non-rigid'' to describe more general deformation patterns beyond articulated movement.

The primary objective in motion representation is to establish accurate correspondences of 3D points across successive temporal frames. Formally, given a point \(\mathbf{x}_{t-1} \in \mathbb{R}^3\) at time \(t-1\), its position \(\mathbf{x}_t\) at time \(t\) can be described by:
\begin{equation}
    \mathbf{x}_{t} 
    \;=\; 
    \mathcal{T}_{\theta}\bigl(\mathbf{x}_{t-1} ; \pi\bigr(t)),
\end{equation}
where \(\mathcal{T}_{\theta}(\cdot)\) is a transformation function parameterized by \(\theta\) and conditioned on temporal context \(\pi\) (e.g., time $t$, frame index $i$, latent code $\ell_t$, or motion-specific parameters). The precise form of \(\mathcal{T}_{\theta}(\cdot)\) depends on the underlying motion pattern and representation method, typically involving different assumptions and formulations.

\subsection{Motion Types}
\label{sec:motion_types}

\subsubsection{Rigid Motion}

Rigid motion encompasses transformations in which an object preserves its shape and size while undergoing rotation and translation. Internally, distances and angles remain unchanged, and thus no deformation occurs. This can be mathematically expressed via a rotation matrix \(\mathbf{R} \in SO(3)\) (or quaternion \(\mathbf{q} \in \mathbb{R}^{4}\)) and a translation vector \(\mathbf{t} \in \mathbb{R}^3\):
\begin{equation}
    \mathbf{x}_{t}
    \;=\;
    \mathcal{T}_{\theta}\bigl(\mathbf{x}_{t-1}\bigr)
    \;=\;
    \mathbf{R}\,\mathbf{x}_{t-1}
    \;+\;
    \mathbf{t}.
\end{equation}
Rigid objects are ubiquitous in daily life, including furniture like boxes and chairs, kitchenware, and other manufactured items. Additionally, certain objects such as vehicles, though not strictly rigid, can be effectively approximated as rigid bodies for many applications, as they maintain a relatively fixed internal structure while translating or rotating \cite{heNeuralRadianceFieldAutonomousDriving2024}. In dynamic view synthesis, accurately tracking and updating an object's rigid 6DoF pose  across frames ensures correct rendering from novel viewpoints~\cite{turki2023suds, fischer2024dynamic, chenOmniReOmniUrbanSceneReconstruction2024a}.

\subsubsection{Articulated Motion}

Articulated motion, also known as piecewise rigid motion, describes a class of transformations where individual segments (e.g., limbs in a skeletal model) undergo rigid transformations while the object's overall motion appears non-rigid due to the relative movement between segments. This type of motion exists in both natural organisms (human and animal skeletons) and manufactured systems (robotic arms, mechanical assemblies, and hinged objects).

An articulated object is typically represented by a kinematic tree--a hierarchical structure of rigid segments connected by joints that provide specific degrees of freedom (e.g., rotation or translation). This hierarchical organization captures how local transformations at each joint propagate through the kinematic chain. The human body represents a quintessential example of such articulated structures, and various parameterization strategies, such as SCAPE \cite{allen2003space}, SMPL \cite{loper2015smpl}, SMPL-X \cite{pavlakos2019expressive}, and MANO \cite{romero2017embodied},  have produced influential frameworks for modeling its complex motion, among which SMPL has gained particular prominence due to its balance of expressiveness and computational efficiency. SMPL conceptualizes the human body as a kinematic tree with 24 joints, each undergoing a rigid transformation relative to its parent, with the root joint defining the transformation from body space to world space. The model employs linear blend skinning (LBS) to deform a canonical rest pose according to a target pose configuration. By assigning blend weights to each vertex on a canonical mesh, SMPL effectively encodes how much each joint's transformation influences that point. Formally, for a point \(\mathbf{x}^c_i\) in the canonical space, its deformed position \(\mathbf{x}^p_i\) in the posed space is computed as follows:
\begin{equation}
    \mathbf{x}^p_i 
    \;=\; 
    \mathcal{T}_{\theta}\!\bigl(\mathbf{x}^c_i,\mathbf{w}(\mathbf{x}^c_i);\pi(J)\bigr)
    \;=\; 
    \sum_{j=1}^{J} 
    \bigl(w_j\!\bigl(\mathbf{x}^c_i\bigr)\,\textbf{T}_j\,\bigr)\mathbf{x}^c_i, 
    \label{eq:lbs}
\end{equation}
where \(\pi(J)\) represents the pose parameters, \(w_j(\mathbf{x}^c_i)\) are blend weights that determine the influence of \(j\) th joint on point \(\mathbf{x}^c_i\), and \(\textbf{T}_j\in \mathbb{R}^{4\times 4}\) is the rigid transformation of the \(j\)-th joint. 

While LBS accurately describes transformations for points on the template surface, handling points in free space (e.g., for volumetric rendering) requires additional techniques. Traditional approaches use barycentric interpolation or nearest-neighbor methods to extend blend weights to points outside the surface~\cite{peng2021neural,liu2021neural,su2023npc}. However, these approaches may struggle with complex deformations, especially for loose clothing or accessories. To address these limitations, recent approaches employ neural skinning fields that replace or augment traditional pre-defined blend weights. A learnable skinning function maps a point \(\mathbf{x}\) to its blending weights:
\begin{equation}
    \mathbf{w}(\mathbf{x}) 
    \;=\; 
    {\mathcal{S}}_\theta\!\bigl(\mathbf{x}; \pi(t)\bigr), 
\end{equation}
where \(\pi(t)\) encodes time-dependent conditions, e.g. pose parameters. Learning such fields from scratch presents challenges; therefore, many methods incorporate the predefined SMPL blend weights as a strong prior, allowing the learned skinning weights to deviate moderately from the canonical configuration when necessary~\cite{weng2022humannerf,li2022tava}.

\subsubsection{Non-Rigid Motion}
Non-rigid motion permits objects to deform by locally changing the relative positions of points and is therefore indispensable for describing cloth folding, facial expression, fluid flow, and other complex dynamics. Unlike rigid or articulated motion, which can exploit kinematic templates or low-dimensional parametric models, general non-rigid deformation typically lacks structured representations due to potential topology changes and the infinite degrees of freedom. Classical tracking pipelines rely on sparse feature matching, yielding motion fields that are far too coarse to capture fine-scale deformation. Physically based methods may provide accurate solutions, yet their computational cost becomes prohibitive for large scenes or real-time use.

Recent advances have reformulated non-rigid motion as an implicit, learnable neural field. These approaches employ compact neural networks to predict dense, continuous displacement fields of the form:
\begin{equation}
    \mathbf{x}_t = \mathcal{T}_{\theta}\bigl(\mathbf{x}_{t-1};\pi(t)\bigr)
    \;=\;
    \mathbf{x}_{t-1}
    \;+\;
    \Delta_{\theta}\!\bigl(\mathbf{x}_{t-1};\pi(t)\bigr).
\end{equation}
where \(\Delta_{\theta}\) predicts the point displacement conditioned on a temporal code \(\pi(t)\).  Such motion fields can be trained directly from 2D images through differentiable rendering, dispensing with explicit 3D supervision and delivering per-point motion estimates. However, this learning problem is normally ill-posed--especially under limited observations,so auxiliary information and extra regularization are often introduced for reasonable solution (Sec. \ref{sec.auxi}).

\subsubsection{Hybrid Motion}
Most real-world scenes exhibit a mixture of motion types--rigid, articulated, and general non-rigid--whose simultaneous presence gives rise to hybrid motion. A prime example is the human body: a primarily articulated skeleton undergoes a global rigid transformation, while soft tissue, loose clothing, and hair introduce finer, highly non-rigid dynamics~\cite{qiu2023rec,jiang2022selfrecon}. Capturing this interplay demands a representation that is both structured enough to model global motion and flexible enough to accommodate local deviations.

Contemporary approaches typically factorize the total motion into complementary components:
\begin{equation}
    \mathbf{x}_t
    \;=\;
    \underbrace{\mathcal{T}_{\theta_1}\bigl(\mathbf{x}_{t-1};\pi(t)\bigr)}_{\text{coarse, e.g.\ rigid/articulated}}
    \;+\;
    \underbrace{\Delta_{\theta_2}\bigl(\mathbf{x}_{t-1};\pi(t)\bigr)}_{\text{fine, e.g.\ non-rigid residual}}.
\end{equation}
In this formulation, $\mathcal{T}_{\theta_1}$ represents a relatively constrained, interpretable global motion model (e.g., rigid transformation or articulated skeleton deformation), while $\Delta_{\theta_2}$ is typically implemented as a neural field that predicts residual displacements to capture complex, local deformations. Besides effective human and animal modeling, similar approaches have been applied to other domains with composite motion, including deformable objects with near-rigid parts and multi-object scenes with varying motion characteristics~\cite{chenOmniReOmniUrbanSceneReconstruction2024a,fischer2024dynamic}. This hierarchical decomposition offers several advantages: it maintains physical interpretability by isolating well-defined transformations; it reduces the complexity of the learning problem by having the neural component focus on residual details rather than the entire motion; and it provides explicit control over coarse motion while allowing the capture of fine details.

\subsection{Motion Representation}
\label{sec:motion_rep}

\subsubsection{Representing via 4D Spacetime}
Starting from coordinate-based representations for static scenes~\cite{adelson1991plenoptic,mildenhall2021nerf}, which utilize the 5D plenoptic function to represent radiance fields as \(\mathcal{F}_\theta\colon (\mathbf{x}, \mathbf{d}) \longmapsto (\mathbf{c},\sigma)\), approaches for dynamic scenes naturally extend the input domain to 6D by incorporating temporal information \(t\)~\cite{xian2021space,parkTemporalInterpolationAllYouNeed2023,liu2023robust,songNerfplayerStreamableDynamicSceneRepresentation2023a}, resulting in \(\mathcal{F}_\theta\colon (\mathbf{x},\mathbf{d},t) \longmapsto (\mathbf{c},\sigma)\). More generally, methods may condition on alternative temporal encodings $\pi(t)$ beyond direct time input, such as frame indices \(i\) or learnable per-frame latent codes \(\ell_i\)~\cite{lombardiNeuralVolumesLearningDynamicRenderable2019,li2022neural,wu2NeRFSelfSupervisedDecouplingDynamicStatic2022}.

With this formulation, each frame of the dynamic scene is optimized independently through frame-by-frame optimization, leveraging dense observations to minimize a rendering loss between the synthesized output \(I_r\) and the ground-truth image \(I_{gt}\):
\begin{equation}
    \arg\min_{\theta} \sum_{} \left\| I_r - I_{gt}\right\|_2^2,\ \ I_r 
    = 
    \mathcal{R}\Bigl(\mathcal{F}_{\theta}\bigl(\mathbf{x}, \mathbf{d}, \pi(t)\bigr)\Bigr),
    \label{eq:optimization_problem}
\end{equation}

where \(\mathcal{R}\) represents the differentiable volume rendering or splatting function. Motion is thus implicitly encoded within the same radiance field that represents the scene--rather than being handled by a separate motion field--and is supervised solely through available 2D image observations.

\subsubsection{Representing via Canonical Space}

A prevalent approach for modeling dynamic scenes decomposes scenes into a static canonical space and time-varying deformation fields. This canonical space--often designated as a "reference frame"--serves as a common coordinate system from which all observed frames are derived through learned deformations. For rigid or articulated objects, this canonical space could be intuitively defined, e.g., initial state for rigid objects or neutral pose for articulated objects. For general scenes, it should capture sufficient geometric detail to facilitate robust correspondence estimation across the sequence.

The relationship between the canonical space and each observation frame is formalized through deformation fields implemented as neural networks. A forward deformation field $\Phi_\theta$ maps points from canonical space to the observation space at time $t$:
\begin{equation}
    \Delta_{c \to i}(\mathbf{x}_c)
    \;=\; 
    \Phi_\theta\bigl(\mathbf{x}_c;\ \pi(t)\bigr),
\end{equation}
Conversely, a backward deformation field $\Psi_\theta$ maps observation space points back to canonical space:
\begin{equation}
    \Delta_{i \to c}(\mathbf{x}_i)
    \;=\; 
    \Psi_\theta\bigl(\mathbf{x}_i;\ \pi(t)\bigr).
\end{equation}
These fields should ideally satisfy invertibility, where $\Phi_\theta \equiv \Psi_\theta^{-1}$, ensuring consistent bidirectional mappings between spaces. This constraint is typically enforced through inverse-consistency regularization or explicitly using invertible neural architectures~\cite{liuMoDGSDynamicGaussianSplattingCausuallycaptured2024b,wangTrackingEverythingEverywhereAllOnce2023,caiNeuralSurfaceReconstructionDynamicScenes2022}.

Within this framework, point correspondence between any two observation frames $i$ and $j$ could be set up by taking the shared canonical space as a intermediate station:
\begin{equation}
    \Delta_{i \to j} 
    \;=\; 
    \Delta_{i \to c} + \Delta_{c \to j},
    \label{cano_conn}
\end{equation}

\subsubsection{Representing via Flow Field} \label{sec:scene_flow}

A more direct strategy for modeling dynamic scenes involves capturing motion between consecutive frames rather than relating observations to a shared reference. This frame-to-frame approach leverages incremental deformations to represent complex motions, decomposing significant transformations into smaller, more tractable steps. Such representations naturally accommodate topology changes and extreme deformations that challenge canonical space methods.

The flow field provides a principled framework for modeling these consecutive-frame dynamics. In their continuous form, velocity fields~\cite{niemeyer2019occupancy,zhou2024hugs,wangFlowSupervisionDeformableNerf2023} specify instantaneous motion by assigning a velocity vector $\mathbf{v}(\mathbf{x}, t)$ to each point $\mathbf{x}$ in space at time $t$. The point movement of point $\textbf{x}$ from time $t-1$ to $t$ can be represented by integrating this velocity field:
\begin{equation}
    \mathbf{x}_t 
    \;=\; 
    \mathbf{x}_{t-1}
    \;+\;
    \int_{t-1}^{t} \mathbf{v}\bigl(\mathbf{x}(\tau), \tau\bigr)\, d\tau,
\end{equation}
where $\mathbf{v}(\mathbf{x}(\tau), \tau)$ represents the velocity at intermediate time $\tau$. While theoretically elegant, obtaining continuous ground-truth velocities is often practically infeasible.

For discrete time steps, scene flow fields $\mathbf{O} (\mathbf{x}, t)$ directly model displacement vectors between consecutive frames:
\begin{equation}
    \mathbf{O}(\mathbf{x}, t) = \Delta_t
    \;=\;
    \mathbf{x}_t 
    \;-\;
    \mathbf{x}_{t-1}.
\end{equation}
Scene flow and velocity fields are mathematically related--velocity represents the time derivative of scene flow--expressed differentially as:
\begin{equation}
    \frac{\partial \mathbf{O}}{\partial t} (\mathbf{x}, t)
    \;=\;
    \mathbf{v}\bigl(\mathbf{O}(\mathbf{x}, t),\, t\bigr),
    \quad
    \text{s.t.}
    \quad
    \mathbf{O}(\mathbf{x},\, t-1) = \mathbf{x}_{t-1}.
\end{equation}
In practical implementations, both fields are typically parameterized using neural networks as $\mathbf{v}_{\theta}$ and $\mathbf{O}_{\theta}$  and optimized with scene radiance field through differentiable rendering. %

These local flow fields can be composed to represent extended movements between far-away frames through:
\begin{equation}
    \Delta_{i \to k}
    \;=\;
    \Delta_{i \to j}
    \;+\;
    \Delta_{j \to k},
    \label{flow_conn}
\end{equation}
with inverse mappings defined as $\Delta_{j \to i} \;\equiv \; \Delta_{i \to j}^{-1}$.

\subsubsection{Representing via Point Tracking}

Recovering motion ultimately reduces to establishing reliable point correspondences across time. Early solutions include sparse feature matching and optical flow. Sparse feature matching identifies distinctive keypoints that can be matched across views and has fueled SfM and SLAM pipelines~\cite{rublee2011orb}. While effective for visual localization, these sparse correspondences reveal little about dense, non-rigid dynamics. Optical flow extends this by estimating dense 2D correspondences between successive frames~\cite{lucas1981iterative,sun2018pwc,teed2020raft}, but long-range tracking quickly deteriorates under appearance changes, occlusions, or large viewpoint shifts. several works chain short-term matches into longer 2D point trajectories~\cite{sand2008particle,doersch2022tap,harley2022particle}. However, purely image-plane tracking struggles with out-of-plane motions that are more naturally handled in 3D. 

Recent progress in differentiable rendering and radiance field has enabled dense, long-term 3D tracking driven by only 2D supervision~\cite{wang2021neural,wangTrackingEverythingEverywhereAllOnce2023,luiten2024dynamic}. These approaches jointly optimize a scene volume and a continuous trajectory field:
\begin{equation}
    \mathbf{x}_t = \mathcal{J}_\theta(t),
\end{equation}
where $\mathcal{J}_\theta$ describes 3D trajectory of any point in 3D space. Rather than stitching local matches via Eq.~\ref{cano_conn} or Eq.~\ref{flow_conn}, fitting such per-point trajectory $\mathcal{J}_\theta$ directly over the full sequence as a whole enforces temporal consistency~\cite{chen2023periodic,stearns2024dynamic}.

\subsubsection{Representing via Factorization}
Decomposing high-dimensional, complex signals into lower-dimensional, simpler components represents a fundamental strategy for signal processing, which is also applicable in motion analysis. This approach significantly reduces computational complexity while preserving essential motion characteristics. For static neural radiance fields, methods such as TensoRF~\cite{chen2022tensorf} and EG3D~\cite{chan2022efficient} have demonstrated the effectiveness of this principle by representing 3D volume via 2D tensors or triplanes, achieving both efficient training and rendering.  

When extending to dynamic scenes, the 4D spacetime introduces additional complexity that can be effectively managed through factorization techniques. Typically, this 4D domain could be decomposed into separable components: static spatial features captured in $({\mathbf{f}_{xy}}, {\mathbf{f}_{yz}}, {\mathbf{f}_{xz}})$  planes, and motion-related temporal components represented in $({\mathbf{f}_{xt}}, {\mathbf{f}_{yt}}, {\mathbf{f}_{zt}})$ planes. For any 3D point $\mathbf{x}$ at a specific time $t$, its features are interpolated from these orthogonal hyperplanes and processed by a neural network to predict the radiance field properties:
\begin{equation}
    \mathcal{F}_\theta(\mathbf{f}_{xy,yz,zx}(\mathbf{x}),\mathbf{f}_{xt,yt,zt}(\mathbf{x})) \longmapsto  (\mathbf{c}, \sigma).
\end{equation}
This hyperplane-based factorization effectively disentangles spatial and temporal components, enabling more efficient optimization and better generalization.  

For scenarios where motion is modeled separately, such as in deformation fields or flow fields, a basis-driven decomposition provides an elegant solution. This approach leverages a limited set of shared motion bases $\{\mathbf{b}_i\}_{i=1}^B$ and time-dependent coefficients $c_i(t)$ to represent each point's motion trajectory:  
\begin{equation}
    \mathbf{x}(t) = \sum_{i=1}^{B} c_i(t) \, \mathbf{b}_i, \quad \text{with} \ \mathbf{b}_i \in \mathbb{R}^3, \ c_i(t) \in \mathbb{R}
\end{equation}
This formulation is particularly powerful because in real-world scenes, nearby points often share similar motion patterns. Since the number of basis functions $B$ is typically much smaller than the number of points in the scene, this creates a well-constrained factorization problem that enhances regularization and temporal consistency.  

The shared basis vectors $\mathbf{b}_i$ can be implemented as learnable parameters or derived from orthogonal function families such as Fourier or sinusoidal expansions. The coefficients $c_i(t)$ dynamically weight each basis vector's contribution to a point's overall movement. Both the basis vectors and coefficients can be jointly optimized with scene geometry and appearance, creating a compact motion representation that maintains global consistency over time.

\begin{figure}
    \centering
    \includegraphics[width=1.0\linewidth]{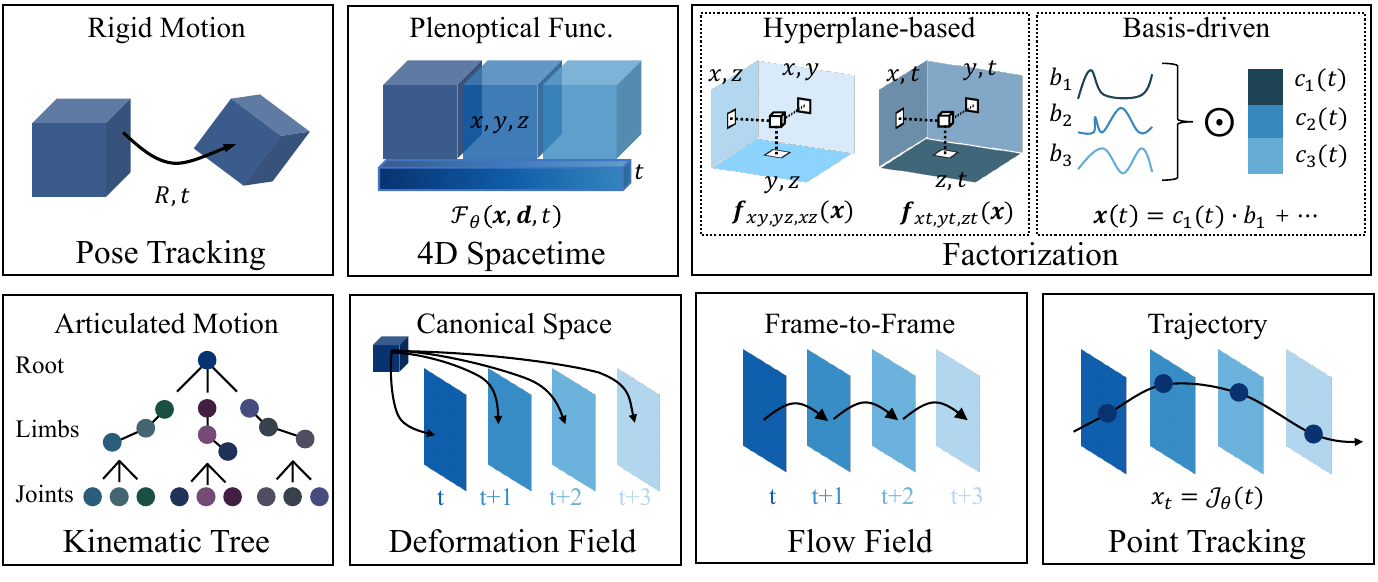}
    \caption{Illustration of typical motion representation methods.}
    \label{fig:motion_repre}
\end{figure}

\begin{table*}
\centering
\begin{tabular}{@{}lcccccccc@{}}
\toprule
\multirow{2}{*}[-1.ex]{Method}      &\multirow{2}{*}[-1.ex]{Venue}       & \multirow{2}{*}[-1.ex]{Input}           & \multicolumn{4}{c}{Auxilary}     & \multirow{2}{*}[-1.ex]{Motion rep.}     & \multirow{2}{*}[-1.ex]{Scene rep.}      \\ \cmidrule(l){4-7}
                                            &     &     &Lidar  &Depth  & Seg.  & O.F.        &                     &                  \\ \midrule
NSG~\cite{ostNeuralSceneGraphsDynamicScenes2021}     &  CVPR'21   & stereo                 &          &     &              &         & scene graph           & MLP           \\
STaR~\cite{yuan2021star}             &  CVPR'21      & multi-view              &     &       &         &            & learnable pose               & MLP  \\

PNF~\cite{kundu2022panoptic}    & CVPR'22   & monocular         &     &               & \cmark        &          & pose tracking             & MLP \\

Mars~\cite{wu2023mars}     & CICIA'23   & monocular         &     & \cmark              & \cmark        &         & pose tracking            & MLP/voxel \\ 
UniSim~\cite{yang2023unisim}           &    CVPR'23         & monocular              & \cmark    &               &         &          & scene graph                 & feature grids \\
S-NeRF~\cite{citation-0}        &   ICLR'23      & multi-view              & \cmark     &              & \cmark        & \cmark           & learnable pose               & MLP \\

ML NSG~\cite{fischerMultilevelNeuralSceneGraphsDynamic2024}        &   CVPR'24            & monocular         & \cmark    &               &         &            & scene graph            & MLP  \\
HUGS~\cite{zhou2024hugs}          &   CVPR'24           & monocular   &     &        & \cmark        & \cmark            & unicycle model                & 3DGS \\

SG~\cite{yan2024street}           &  ECCV'24      & monocular              & \cmark    &               &         &             & learnable pose                 & 3DGS  \\ 
 
NeuRAD~\cite{tonderski2024neurad}       &   CVPR'24          & monocular         & \cmark    &          &         &            & pose tracking             & MLP \\ 

DrivingGaussian~\cite{zhou2024drivinggaussian}      &  CVPR'24       & multi-view              & \cmark     &        & \cmark        &         & scene graph             & 3DGS \\

AutoSplat~\cite{khan2024autosplat}        &      ICRA'25          & monocular         & \cmark    &               & \cmark        &            & pose tracking             & 3DGS \\
\bottomrule 
\end{tabular}
\caption{
Selected papers for dynamic scene reconstruction with rigid motion. 
}
\label{tab:rigid}
\end{table*}

\subsection{Disscussion}
\label{sec:disscuss}

Dynamic motion representation methods span a spectrum balancing structural constraints and flexibility, with each approach offering distinct trade-offs, illustrated in Fig.~\ref{fig:motion_repre}. Rigid motion representations efficiently model objects through 6DoF pose tracking (sometimes incorporating scale factors for distance variations), providing computational efficiency and interpretability while inherently limiting complex deformations. Articulated motion extends this concept by modeling hierarchical relationships between connected rigid parts through kinematic chains, effectively representing entities like human bodies while typically requiring category-specific templates as prior knowledge. Canonical space with deformation fields elegantly disentangles geometry and appearance from motion by decomposing scenes into a static canonical space and time-varying deformation fields, enabling sophisticated motion analysis and canonical manipulation, though defining appropriate canonical spaces becomes challenging for sequences with large deformations or extended temporal spans.  

4D spacetime representations directly extend static scene modeling to the temporal domain for frame-by-frame optimization, offering conceptual simplicity but lacking explicit point correspondence and struggling with cross-view consistency. Frame-to-frame flow fields model incremental deformations between consecutive frames, naturally accommodating topology changes without requiring global canonical spaces, though they accumulate errors over extended sequences. Point tracking approaches represent trajectories as continuous functions across time, establishing dense 3D correspondences while remaining vulnerable to occlusions. Factorization-based representations decompose motion into shared basis functions and time-dependent coefficients, significantly reducing parameter dimensionality while enforcing motion coherence between spatially proximate points. This representation spectrum reflects fundamental trade-offs: rigid and articulated approaches impose strong priors but sacrifice adaptability, while frame-by-frame methods maximize flexibility at the cost of temporal consistency, suggesting that hybrid approaches combining complementary paradigms often yield superior results for complex real-world applications.

\section{Dynamic Scene Reconstruction}
\label{sec:scene_repre}
%
\subsection{Reconstructing with Rigid Motion}

Rigid motion reconstruction fundamentally revolves around pose tracking, estimating the 6DoF transformation of each object over time. By tracking the 3D bounding boxes of moving objects, these methods spatially decomposes dynamic scenes into foreground objects and static backgrounds~\cite{yan2024street}, with some methods incorporating dedicated sky modules to handle distant, uncertain regions~\cite{wu2023mars}. Early works such as NSG~\cite{ostNeuralSceneGraphsDynamicScenes2021} introduced the scene graph, a hierarchical structure where nodes represent individual scene elements (objects or background) and edges encode their spatial relationships as rigid transformation, enabling efficient novel view synthesis of complex dynamic scenes. This concept evolved into multi-level scene graph in ML NSG~\cite{fischerMultilevelNeuralSceneGraphsDynamic2024} and dynamic scene graph in ProSGNeRF~\cite{dengProsgnerfProgressiveDynamicNeuralScene2023}. Multi-view systems like STaR~\cite{yuan2021star} employ self-supervised tracking alongside scene graph to jointly optimize the object poses without manual annotations.

Temporal coherence is achieved by linking object instances across frames through consistent pose tracking. Methods like PNF~\cite{kundu2022panoptic} additionally employ meta-learning techniques to initialize category-specific object fields, while HUGS~\cite{zhou2024hugs} constrains vehicles to ground-plane motion using a unicycle model for improved stability. More recent approaches, such as S-NeRF~\cite{citation-0} and SG~\cite{yan2024street}, optimize tracked poses jointly with scene parameters, yielding more accurate reconstruction and alignment across frames. 

A recent trend is the transition from implicit neural radiance fields (parameterized by MLPs) to explicit 3DGS representations for faster convergence and real-time rendering capabilities. To constrain the ill-posed nature of dynamic scene reconstruction (especially from monocular inputs), methods often leverage auxiliary information such as sementic segmentation~\cite{kundu2022panoptic,zhou2024hugs,wu2023mars,zhou2024drivinggaussian,khan2024autosplat}, optical flow~\cite{zhou2024hugs,citation-0}, or Lidar~\cite{tonderski2024neurad,khan2024autosplat} and depth data~\cite{wu2023mars} to offer precise boundary, constrain motion space, and resolve scale ambiguity. UniSim~\cite{yang2023unisim} further incorporate feature grids to model finer environmental details while maintaining efficient object-level motion representation. The practical applications of these approaches span autonomous driving, augmented reality, and robotics. Explicit scene decomposition through rigid motion tracking enables capabilities like object removal~\cite{ostNeuralSceneGraphsDynamicScenes2021}, trajectory modification~\cite{tonderski2024neurad}, and viewpoint manipulation~\cite{yang2023unisim}, all crucial for simulation environments and digital twins.

\subsection{Reconstructing with Articulated Motion}

\begin{table*}
\centering
\resizebox{0.88\textwidth}{!}{
\begin{tabular}{@{}llccccccc@{}}
\toprule
\multirow{2}{*}[-1.ex]{Method}      &\multirow{2}{*}[-1.ex]{Venue}       & \multirow{2}{*}[-1.ex]{Input}           & \multicolumn{4}{c}{Auxilary}     & \multirow{2}{*}[-1.ex]{Motion rep.}     & \multirow{2}{*}[-1.ex]{Obj. rep.}       \\ \cmidrule(l){4-7} 
                            &    &      &Prior  &Norm  &Mask & L.P.  &   &    \\ 
\midrule \multicolumn{9}{l}{Human body} \\ \midrule 
Neural Body~\cite{peng2021neural}     & CVPR'21    &multi-view   &SMPL &    & &     &forward skinning    &voxel grids     \\
A-NeRF~\cite{su2021nerf}      & NeurIPS'21  &monocular         &skeleton    &    & &\cmark    &skeleton-relative encoding        &MLP  \\
NARF~\cite{noguchi2021neural} &ICCV'21 &monocular  &skeleton &  &\cmark & &forward skinning  &part-wise     \\
AN~\cite{peng2021animatable}&ICCV'21 &multi-view  &SMPL &  & & &invserse skinning  & MLP     \\
NHP~\cite{kwon2021neural}&NeurIPS'21 &multi-view  &SMPL &  & & &invserse skinning  & p.a. feat.     \\
Vid2Avatar~\cite{guo2023vid2avatar} &CVPR'23 &casual  &skeleton &\cmark &\cmark & &  inverse skinning & MLP \\
MonoHuman~\cite{yu2023monohuman} &CVPR'23 &monocular  &SMPL &  &\cmark & &bidirectional deformation  & MLP     \\
InstantAvatar~\cite{yu2023monohuman} &CVPR'23 &monocular  &skeleton &  &\cmark & &forward skinning  & HashGrids     \\
ExAvatar~\cite{moon2024expressive}  &ECCV'24    & casual   &SMPL-X     & \cmark  & \cmark        &\cmark         &forward skinning    & 3DGS \\ 
HuGS~\cite{kocabas2024hugs} &CVPR'24 &monocular  &SMPL &  &\cmark & &forward skinning  & 3DGS     \\
GART~\cite{lei2024gart} &CVPR'24 &monocular  &SMPL &  & & &forward skinning  & 3DGS     \\
GauHuman~\cite{hu2024gauhuman}&CVPR'24 &monocular  &SMPL &  & &\cmark &forward skinning  & 3DGS     \\
Animatable Gaussians!\cite{li2024animatable} &CVPR'24 &multi-view  &SMPL-X &  & & &forward skinning  & 3DGS     \\
GaussianAvatar\cite{hu2024gaussianavatar} &CVPR'24 &monocular  &SMPL-X &  & &\cmark &forward skinning  & 3DGS     \\
ASH~\cite{pang2024ash} &CVPR'24 &multi-view  &skeleton &  & & &DQS  & 3DGS     \\
MoDA~\cite{song2024moda} &IJCV'24 &casual  & &  &\cmark & &NeuDBS  &MLP     \\

\midrule \multicolumn{9}{l}{Hand} \\ \midrule 
LISA~\cite{corona2022lisa}  &CVPR'22 &multi-view  &MANO &  & &\cmark &inverse skinning  &SDF     \\
HandAvatar~\cite{chen2023hand}  &CVPR'23 &monocular  &MANO-HD &  & & &forward skinning  &Occupancy     \\
LiveHand~\cite{mundra2023livehand}  &ICCV'23 &multi-view  &MANO &  & & &UVH  &MLP     \\
GaussianHand~\cite{zhao2024gaussianhand}  &TVCG'24 &multi-view  &MANO &  & & &forward skinning  &3DGS     \\
MANUS~\cite{pokhariya2024manus}  &CVPR'24 &multi-view  &skeleton &  & & &forward skinning  &3DGS     \\

\midrule \multicolumn{9}{l}{Animal} \\ \midrule 
ARTEMIS~\cite{luo2022artemis}  &TOG'22 &multi-view  &skeleton &  &\cmark & &forward skinning  &voxel grids     \\ 
BANMo~\cite{yang2022banmo}   &CVPR'22 &casual  & &  &\cmark &\cmark   &forward skinning  &MLP     \\ 
MagicPony~\cite{wu2023magicpony}   &CVPR'23 &single view  &skeleton &  &\cmark &   &forward skinning  &SDF     \\ 
CoP3D~\cite{sinha2023common}   &CVPR'23 &casual  & &  &\cmark &   &trajectory  &p.a. feat.     \\ 
AnimalAvatar~\cite{sabathier2024animal}   &ECCV'24 &monocular  &SMAL &  &\cmark &\cmark   &forward skinning  &Triplane     \\

\midrule \multicolumn{9}{l}{Object} \\ \midrule 
CLA-NeRF~\cite{tseng2022cla}   &ICRA'22 &multi-view  & &  &\cmark &\cmark   &joint para.  &MLP     \\ 
PARIS~\cite{liu2023paris}   &ICCV'23 &multi-view  & &  &\cmark &   &joint para.  &HashGrids     \\ 
LEIA~\cite{swaminathan2024leia}   &ECCV'24 &multi-view  & &  & &   &state code  &MLP     \\
REACTO~\cite{song2024reacto}   &CVPR'24 &casual  & &  &\cmark &   &QRBS  &MLP     \\
ArtGS~\cite{liu2025artgs}   &ICLR'25 &multi-view  & &  & &   &SE(3)  &3DGS     \\

\bottomrule 
\end{tabular}
}
\caption{
Selected papers for dynamic scene reconstruction with articulated motion. L.P. stands for \textbf{L}earnable \textbf{P}ose.
}
\label{tab:articulated}
\end{table*}

\textbf{Human Body.} %
Articulated human body reconstruction has evolved through distinct developmental stages, progressing from simple 2D/3D pose estimation and basic mesh recovery~\cite{liuDeepLearning3DHumanPose2024} toward photorealistic appearance and fine geometric detail reconstruction. This advancement has been particularly accelerated by the introduction of implicit neural fields~\cite{park2019deepsdf,mescheder2019occupancy}, especially recent radiance fields~\cite{mildenhall2021nerf,kerbl3Dgaussians}. Early reconstruction approaches strategically integrated parametric human models such as SMPL or SMPL-X as structural priors with neural radiance fields to represent time-varying surface details. These methods established the canonical space paradigm, where the human body is represented in a rest pose while points from the observation space (posed body) are transformed to this canonical reference via inverse skinning to query field properties like color and density values~\cite{guo2023vid2avatar,peng2021animatable,kwon2021neural,weng2022humannerf,jiang2022neuman}.

The learning process is supervised using multi-view or monocular images, with the optimization accounting for both appearance and geometry. A fundamental challenge is diffusing skinning weights to arbitrary points in 3D space. Solutions range from nearest-neighbor interpolation~\cite{noguchi2021neural,zheng2022structured,su2023npc} and barycentric mapping~\cite{liu2021neural} to more sophisticated approaches like Neural Body's 3D convolutional networks and learnable continuous skinning fields~\cite{li2022tava,jiang2023instant,jiang2023instantavatar}. While early methods defined skinning fields in observation space, these approaches frequently struggle with generalization to novel poses~\cite{chen2021snarf,chen2023fast}. Recent advances define the skinning field in canonical space and employ root-finding algorithms to establish bidirectional point correspondences~\cite{dong2022pina,jiang2023instantavatar,shen2023x,yu2023monohuman}, enabling forward skinning transformations that generalize significantly better to out-of-distribution poses and providing more robust reconstruction across complex articulations. Complementary techniques include pixel-aligned features for cross-identity generalization~\cite{kwon2021neural,cheng2022generalizable,raj2021pixel,yu2021pixelnerf}, part-wise representations for enhanced detail~\cite{shen2023x}, and hybrid representation for more efficient training and rendering~\cite{jiang2023instantavatar,xu20244k4d,habermann2021real}.  

3DGS has emerged as a transformative explicit representation for articulated human bodies, offering both quality improvements and dramatic efficiency gains over implicit neural fields~\cite{moon2024expressive,kocabas2024hugs,lei2024gart,hu2024gauhuman,pang2024ash,li2024animatable,xu2025animatable,jena2023splatarmor}. In a typical 3DGS pipeline, Gaussians are initialized based on a parametric template in rest pose and then transformed into observation space via forward skinning, which naturally drives both position and orientation parameters. During this skinning process, the color and density of the Gaussians are typically fixed as initialized to ensure better convergence~\cite{qian20243dgs}. This scheme fundamentally resolves the correspondence ambiguities present in inverse skinning used by neural implicit representations. Instead of directly optimizing Gaussian parameters, some methods represent Gaussians with learned embeddings, predicting parameters using embeddings sampled via UV mapping~\cite{li2024animatable,hu2024gaussianavatar,pang2024ash}. As 3D Gaussian fields lack inherent structure, advanced methods bind Gaussians to structured meshes or tetrahedral cages to enhance animation control and spatial coherence~\cite{shao2024splattingavatar,wen2024gomavatar}. Several approaches also jointly optimize pose, skinning weights, and skeleton alongside Gaussian parameters~\cite{zhang2024bags}.  

Beyond canonical space methods, alternative approaches represent articulated bodies directly in observation space or pursue template-free reconstruction. Direct observation space techniques leverage specialized parameterizations such as UV-based coordinates~\cite{geng2023learning,liu2021neural} or skeleton-based local coordinates~\cite{su2021nerf,noguchi2021neural,deng2020nasa}, circumventing the need for explicit canonicalization. Meanwhile, template-free approaches learn articulation parameters, including skeleton structures and skinning weights, entirely from scratch, demonstrating remarkable generalizability across different body types and motion patterns~\cite{yang2021lasr,yang2021viser,song2024moda}. These methods typically incorporate auxiliary supervision from silhouettes, semantic segmentation, optical flow, or leverage data-driven priors from foundation models like DINO features~\cite{oquab2024dinov2} or CSE embeddings~\cite{neverova2020continuous} to establish robust correspondences across frames.

\textbf{Hands.} %
Human hands are another typical articulated structure that plays a crucial role in everyday life. Similar to the SMPL body model, MANO~\cite{romero2017embodied} represents hand geometry with a pre-defined skeleton, blend shapes, and models hand motion using pose parameters and skinning weights. While MANO provides a widely-used parametric foundation, its relatively coarse mesh has led to developments like MANO-HD~\cite{chen2023hand}, which offers high-resolution hand geometry while maintaining compatibility with existing MANO-annotated datasets. For realistic hand appearance, researchers have explored diverse representation strategies including neural fields~\cite{corona2022lisa,zheng2024ohta}, radiance fields~\cite{guo2023handnerf,mundra2023livehand}, and texture maps~\cite{qian2020html,karunratanakul2023harp,chen2024urhand}, with methods like RelightableHands~\cite{iwase2023relightablehands}, HandRT~\cite{kalshetti2025handrt}, and URHand~\cite{chen2024urhand} enabling physically-based relighting through explicit material modeling. Hand reconstruction approaches increasingly address complex interaction scenarios, including hand-to-hand~\cite{guo2023handnerf,lee2023im2hands} and hand-to-object interactions~\cite{ye2022s,tu2023consistent}. Recent advances have introduced novel representations such as LiveHand's~\cite{mundra2023livehand} UVH space parameterization that represents hands relative to the MANO surface without explicit skinning, Nimble's~\cite{li2022nimble} modeling of inner bones and muscles for enhanced biomechanical realism. MANUS~\cite{pokhariya2024manus} effectively represents articulated hands by utilizing 3D Gaussian Splatting in canonical space and employing forward skinning to convert these Gaussians into posed space, where the model is supervised through multi-view sequences to achieve precise shape and appearance reconstruction. GaussianHand~\cite{zhao2024gaussianhand} enhances articulated hand modeling by leveraging canonical features to refine blend shapes derived from parametric models like MANO and implementing neural residual skeletons to capture subtle pose-dependent deformations, resulting in a more accurate representation of hand poses than methods using only standard linear blend skinning.

\begin{table*}
\centering
\resizebox{0.88\textwidth}{!}{
\begin{tabular}{@{}llccccccc@{}}
\toprule
\multirow{2}{*}[-1.ex]{Method}      &\multirow{2}{*}[-1.ex]{Venue}       & \multirow{2}{*}[-1.ex]{Input}           & \multicolumn{4}{c}{Auxilary}     & \multirow{2}{*}[-1.ex]{Motion rep.}     & \multirow{2}{*}[-1.ex]{Obj. rep.}       \\ \cmidrule(l){4-7} 
                            &    &      &Depth  &Mask  &O.F. &Reg.  &   &    \\ 
\midrule \multicolumn{9}{l}{4D Spacetime} \\ \midrule 
Neural Volumes~\cite{lombardiNeuralVolumesLearningDynamicRenderable2019}     & ToG'19    &multi-view   & &\cmark & &TV     &latent code    &RGBA volume    \\
VideoNeRF~\cite{xian2021space}   & CVPR'21    &casual   &\cmark & & &CE     &time    &MLP    \\
DyNeRF~\cite{li2022neural}   & CVPR'22    &multi-view   & & & &     &latent code    &MLP    \\
D\textsuperscript{2}NeRF~\cite{wu2NeRFSelfSupervisedDecouplingDynamicStatic2022} &NeurIPS'22 &monocular   & & & &CE     &latent code    &MLP    \\
TiNeuVox~\cite{fangFastDynamicRadianceFieldsTimeaware2022} &SIGGRAPH Asia'22 &multi-view   & & & &CE     &time    &voxel grids    \\
NeRFPlayer~\cite{songNerfplayerStreamableDynamicSceneRepresentation2023a} &TVCG'23 &monocular   & & & &CE     &time    &voxel grids    \\
SUDS~\cite{turki2023suds} &CVPR'23 &monocular   &\cmark & &\cmark &CE/cycle     &frame index    &HashGrids    \\
MLP Maps~\cite{pengRepresentingVolumetricVideosDynamicMlp2023} &CVPR'23 &multi-view   & & & &     &latent code    &factorized planes    \\
4DGS~\cite{yangreal} &ICLR'24 &monocular  & & & &     &time    &3DGS   \\
STG~\cite{liSpacetimeGaussianFeatureSplattingRealtime2024} &CVPR'24 &multi-view  & & & &     &time    &gaussian feature   \\
GFlow~\cite{wang2024gflow}  &AAAI'25 &monocular  &\cmark & &\cmark &     &time    &3DGS   \\

\midrule \multicolumn{9}{l}{Canonical Space with Deformation Field} \\ \midrule 
Nerfies~\cite{park2021nerfies}  &ICCV'21 &multi-view  & &\cmark & &elastic &latent code  &MLP     \\
D-NeRF~\cite{pumarola2021d}  &CVPR'21 &monocular  & & & & &time  &MLP     \\
HyperNeRF~\cite{park2021hypernerf}   &TOG'21 &casual  & & & & &latent code  &MLP     \\
NDVG~\cite{guoNeuralDeformableVoxelGridFast2022a} &ACCV'22 &monocular  & &\cmark & &CE/TV/$L_1$ &time  &MLP    \\
HyperReel~\cite{attal2023hyperreel}   &CVPR'23 &multi-view  & & & & &velocity  &factorized planes     \\
Deformable 3DGS~\cite{yang2024deformable} &CVPR'24 &monocular  & & & & &time  &3DGS    \\
GA-GS~\cite{lu3dGeometryawareDeformableGaussianSplatting2024} &CVPR'24 &monocular  & & & &$L_1$ &time  &3DGS    \\

\midrule \multicolumn{9}{l}{Frame-to-Frame Flow Field} \\ \midrule 
NeRFlow~\cite{duNeuralRadianceFlow4dView2021}  &ICCV'21 &monocular  & & & &   &scene flow  &MLP     \\ 
DynamicNeRF~\cite{gaoDynamicViewSynthesisDynamicMonocular2021a} &ICCV'21 &monocular  &\cmark &\cmark &\cmark &CE/TV/cycle/$L_1$   &scene flow  &MLP     \\ 
NSFF~\cite{liNeuralSceneFlowFieldsSpacetime2021} &CVPR'21 &multi-view  &\cmark &\cmark &\cmark &cycle/$L_1$   &scene flow  &MLP     \\ 
MonoNeRF~\cite{tian2023mononerf}  &ICCV'23 &monocular  &\cmark &\cmark &\cmark &cycle   &velocity field  &MLP     \\ 
FSDNeRF~\cite{wangFlowSupervisionDeformableNerf2023} &CVPR'23 &monocular  & &\cmark &\cmark &   &velocity field  &MLP     \\ 
DynPoint~\cite{zhou2024dynpoint} &NeurIPS'24 &Monocular  &\cmark &\cmark &\cmark &   &scene flow  &neural points     \\ 

\midrule \multicolumn{9}{l}{Point Tracking} \\ \midrule 
OmniMotion~\cite{wangTrackingEverythingEverywhereAllOnce2023}  &ICCV'23 &monocular  & & &\cmark &cycle/ $L_1$   &bijective mapping  &MLP     \\
DynGS~\cite{luiten2024dynamic} &3DV'24 &multi-view  & &\cmark & &ARAP/isometric   &time  &3DGS     \\
Marbles~\cite{stearns2024dynamic} &SIGGRAPH Asia'24 &casual  & &\cmark & &isometric   &trajectory  &3DGS     \\

\midrule \multicolumn{9}{l}{Factorization} \\ \midrule 
NPGs~\cite{dasNeuralParametricGaussiansMonocularNonrigid2024}  &CVPR'24 &monocular  & &\cmark &\cmark &isometric   &basis  &3DGS     \\ 
FPO~\cite{wangFourierPlenoctreesDynamicRadianceField2022a} &CVPR'22 &multi-view  & &\cmark & &    &basis  &voxel grids     \\
Tensor4D~\cite{shao2023tensor4d}   &CVPR'23 &multi-view  & & & &TV   &feature planes  &MLP     \\ 
Hexplane~\cite{caoHexplaneFastRepresentationDynamicScenes2023}  &CVPR'23 &monocular  & & & &TV    &feature planes  &MLP     \\ 
K-Planes~\cite{fridovich2023k}  &CVPR'23 &monocular  & & & &TV/Laplacian/$L_1$    &feature planes  &MLP     \\
4K4D~\cite{xu20244k4d}   &CVPR'24 &multi-view  & &\cmark & &   &feature planes  &MLP     \\ 
4D GS~\cite{wu20244d} &CVPR'24 &monocular  & & & &TV   &feature planes  &3DGS     \\ 
DeformGS~\cite{duisterhof2023deformgs}  &WAFR'24 &multi-view  & &\cmark & &isometric   &feature planes  &3DGS     \\ 
DynMF~\cite{kratimenos2024dynmf}  &ECCV'24 &monocular  & & & &isometric/$L_1$   &basis  &3DGS    \\

\bottomrule 
\end{tabular}
}
\caption{
Selected papers for dynamic scene reconstruction with non-rigid motion.
}
\label{tab:nrigid}
\end{table*}

\textbf{Animal.} %
Reconstructing articulated animals presents unique challenges due to the vast diversity of species, making it difficult to adapt a single template to accommodate all morphologies. The SMAL model~\cite{zuffi20173d}, a pioneering parametric model primarily for quadruped animals, provides a foundation for subsequent research. Based on this parametric framework, researchers have demonstrated that animal shapes can be accurately fitted using only 2D image inputs~\cite{biggs2020left}. More recently, neural radiance field approaches have enabled learning detailed geometry and appearance~\cite{sabathier2024animal,luo2022artemis,yang2022banmo}, allowing for photorealistic novel view synthesis through volume rendering. These methods typically employ a dual-level representation: a category-level template for general morphology combined with instance-level corrections to capture individual time-varying variations~\cite{yangReconstructingAnimatableCategoriesVideos2023}.  

Several innovative approaches have moved beyond template meshes, instead using only skeletal structures as priors~\cite{luo2022artemis,wu2023magicpony,yao2022lassie}, where the posed animal is driven entirely by skeletal motion. Even more remarkably, some methods operate without any species-specific prior knowledge, learning animal models automatically from raw data~\cite{yang2022banmo,sinha2023common,li2024learning}, demonstrating the versatility of articulated motion representations across diverse morphologies. For example, BANMo~\cite{yang2022banmo} learns neural bones and skinning weights directly from casual videos, representing animals as a neural radiance field in canonical space, with point correspondences established through bidirectional neural skinning operations. ARTEMIS~\cite{luo2022artemis} represents geometry and appearance using neural feature voxel grids, with posed animals warped via skeletal motion and rendered through neural rendering techniques. CoP3D~\cite{sinha2023common} utilizes pixel-aligned features to estimate density and color without explicitly modeling motion. Learning the articulation from casually monocular videos remains an inherently ill-posed problem, these methods also leverage various auxiliary information sources to constrain the solution space, including foreground masks~\cite{sabathier2024animal,luo2022artemis}, optical flow~\cite{yang2022banmo,sinha2023common}, surface normals~\cite{yangReconstructingAnimatableCategoriesVideos2023}, and semantically rich features from foundation models such as CSE~\cite{sabathier2024animal} and DINO~\cite{yao2022lassie}.

\textbf{Objects.} %
Unlike human or animal bodies with consistent skeletal structures, general articulated objects (such as laptops, scissors, and other mechanical devices) present unique reconstruction challenges due to their diverse topologies and joint configurations. For these objects, using pre-defined kinematic trees becomes impractical, as each object type features distinct articulation patterns. The key challenge in reconstructing such general articulated objects involves three interrelated tasks: accurately segmenting the constituent parts, defining appropriate joint motion types (e.g., rotational, prismatic), and estimating precise joint motion parameters for each articulated state~\cite{wei2022self}.  

Recent advances in neural radiance field techniques have enabled significant progress in part-level geometry and appearance reconstruction combined with joint motion parameter estimation~\cite{liu2025artgs,tseng2022cla,liu2023paris}. Several approaches address this challenge from different perspectives. PARIS~\cite{liu2023paris} simplifies the problem by assuming objects contain only one movable part and represents articulated motion through explicit joint motion parameters, storing geometry and appearance in efficient Instant-NGP-style hash grids~\cite{muller2022instant}. Taking a different approach, CLA-NeRF~\cite{tseng2022cla} employs category-level semantic neural radiance fields to segment individual parts and represents the motion between each part and a designated root part through rigid transformations. Rather than directly modeling joint motions, LEIA~\cite{swaminathan2024leia} introduces a more abstract approach using latent state codes to represent different articulation states. REACTO~\cite{song2024reacto} offers a more flexible solution by implementing Quasi-Rigid Blend Skinning (QRBS) to represent articulation motion, learning neural bones and skinning weights directly from casually captured monocular video without requiring explicit part segmentation or predefined articulation models.

\subsection{Reconstructing with Non-rigid Motion}

\textbf{4D Spacetime} %
 is a unified representation that implicitly encodes scene geometry, appearance, and motion within a single radiance field, with properties like density and color varying temporally without explicitly modeling motion vectors~\cite{xian2021space,lombardiNeuralVolumesLearningDynamicRenderable2019}. To enhance efficiency and visual fidelity, recent methods decompose scenes into time-invariant static backgrounds (5D) and time-dependent~\cite{pengRepresentingVolumetricVideosDynamicMlp2023,caiNeuralSurfaceReconstructionDynamicScenes2022} or latent code conditioned~\cite{li2022neural} dynamic foregrounds (6D), blended via learned weights and often guided by segmentation masks~\cite{turki2023suds,wangMaskedSpacetimeHashEncodingEfficient2024a} or self-supervised techniques~\cite{wu2022d}. Implementation approaches have evolved from MLPs to more efficient structures: some methods use 4D neural voxels for accelerated rendering~\cite{parkTemporalInterpolationAllYouNeed2023,wangMixedNeuralVoxelsFastMultiview2023,ganV4dVoxel4dNovelView2023}, while others leverage 3D Gaussian splatting to model 4D spacetime as sequential slices of 3D space with time-dependent Gaussian properties~\cite{yangreal,liSpacetimeGaussianFeatureSplattingRealtime2024}, enabling real-time rendering of complex non-rigid motions.

\textbf{Canonical Space with Deformation Field} %
approach decomposes dynamic scenes into a static reference volume (canonical space) and its temporal evolution pattern (deformation field)~\cite{huang2024sc,park2021nerfies,guoNeuralDeformableVoxelGridFast2022a,liu2022devrf}. In neural radiance field implementations, this approach typically employs backward deformation to transform sampled points along camera rays in the observation space back to canonical space~\cite{park2021nerfies,pumarola2021d,park2021hypernerf}, as their density and color properties are stored in the canonical frame. In contrast, explicit representations like 3D Gaussian fields can directly apply forward deformation, warping each Gaussian primitive from canonical to observation space~\cite{yang2024deformable,liuMoDGSDynamicGaussianSplattingCausuallycaptured2024b}. Neither approach alone guarantees perfect consistency between spaces, leading some methods to implement bijective deformation fields that maintain correspondences in both directions~\cite{caiNeuralSurfaceReconstructionDynamicScenes2022}.  

For extended sequences with substantial motion or appearance changes, a single global canonical space often proves insufficient. In such cases, multiple local canonical spaces (keyframes) shared by temporal subwindows provide a more effective solution~\cite{attal2023hyperreel}. This approach allows nearby frames to reference the same keyframe while temporally distant frames leverage different keyframes, better accommodating dramatic transformations while maintaining local consistency.

\textbf{Frame-to-Frame Flow Field} %
models dynamic motion as point correspondences between consecutive frames, known as frame-to-frame scene flow fields. These fields typically formulate the motion relationship between adjacent frames, where smaller displacements make the motion patterns easier to learn and model. This flow field is generally implemented as a 4D function that maps 3D spatial positions and a 1D time parameter to corresponding displacement vectors~\cite{duNeuralRadianceFlow4dView2021,gaoDynamicViewSynthesisDynamicMonocular2021a,yang2023emernerf}.  

Rather than representing flow in only one direction, bidirectional approaches enhance reconstruction quality. For example, Li et al.~\cite{liNeuralSceneFlowFieldsSpacetime2021} utilize both forward and backward flow fields within the same framework, establishing point correspondences between frames $i$ and $j$. When points from frame $i$ move to frame $j$ along flow field $f_{i\rightarrow j}$, the rendered results should maintain consistency with frame $j$, and vice versa. This bidirectional consistency effectively enables information sharing between adjacent frames, serving as a powerful constraint that enhances learning efficiency~\cite{niemeyer2019occupancy}.  

The framework can be extended by transferring points from a target frame to multiple source frames, allowing information aggregation across temporal neighbors for more generalizable and adaptive reconstructions~\cite{tian2023mononerf,sinha2023common,zhou2024dynpoint}. Since scene flow fields naturally exhibit non-zero values only in dynamic regions, decomposing scenes into static and dynamic components significantly benefits the learning of meaningful flow fields~\cite{tian2023mononerf,gaoDynamicViewSynthesisDynamicMonocular2021a,liNeuralSceneFlowFieldsSpacetime2021,zhou2024dynpoint,yang2023emernerf}.  

Beyond direct flow field learning, an alternative formulation treats flow fields as the integration of velocity fields over time~\cite{niemeyer2019occupancy,duNeuralRadianceFlow4dView2021,wangFlowSupervisionDeformableNerf2023}. While flow fields are typically constrained by smoothness and continuity regularizations, velocity fields offer additional physical constraints and directional information through their vector nature. Li et al.~\cite{liNVFiNeuralVelocityFields3D2024} demonstrate this by incorporating physical laws as supervision through physics-informed neural networks (PINNs), enabling applications like future frame extrapolation, motion transfer, and semantic decomposition. Regardless of whether flow fields or velocity fields are employed, 2D optical flow provides valuable supervision signals for learning~\cite{duNeuralRadianceFlow4dView2021,tian2023mononerf,liNeuralSceneFlowFieldsSpacetime2021,zhou2024dynpoint}, particularly in monocular settings where depth information is limited.

\begin{table*}
\centering
\begin{tabular}{@{}llccccccc@{}}
\toprule
\multirow{2}{*}[-1.ex]{Method}      &\multirow{2}{*}[-1.ex]{Venue}       & \multirow{2}{*}[-1.ex]{Input}           & \multicolumn{4}{c}{Auxilary}     & \multirow{2}{*}[-1.ex]{Motion rep.}     & \multirow{2}{*}[-1.ex]{Obj. rep.}       \\ \cmidrule(l){4-7} 
                            &    &      &Prior  &O.F.  &Mask &Depth  &   &    \\ \midrule
NA~\cite{liu2021neural} &TOG'21   &multi-view         &SMPL     &  &  &      &invserse skinning+deformation          &MLP   \\
NeuMan~\cite{jiang2022neuman}  &ECCV'22   & monocular   &SMPL     &        & \cmark        &             & inverse skinning+deformation     & MLP  \\
HumanNeRF~\cite{weng2022humannerf} &CVPR'22 &casual  &skeleton &  &\cmark &  &inverse skinning+deformation & MLP \\
TAVA~\cite{li2022tava} &ECCV'22 &multi-view  &skeleton &  & & &skinning+deformation & MLP \\

Instant-NVR~\cite{geng2023learning}&CVPR'23 &monocular  &SMPL &  & & &inverse skinning+deformation  &part-wise     \\
HandNeRF~\cite{guo2023handnerf} &CVPR'23 &multi-view  &MANO &  &\cmark & &inverse skinning+deformation  &MLP     \\

HOSNeRF~\cite{liu2023hosnerf} &CVPR'23 &monocular  &skeleton &\cmark  &\cmark & &inverse skinning+deformation  &MLP     \\
ExAvatar~\cite{moon2024expressive}  &ECCV'24    & casual   &SMPL-X     &   & \cmark        &         &skinning+deformation    & 3DGS \\ 

GoMAvatar~\cite{wen2024gomavatar} &CVPR'24 &monocular  &SMPL &  &\cmark & &skinning+deformation & 3DGS \\
3DGS-Avatar~\cite{qian20243dgs} &CVPR'24 &monocular  &SMPL &  &\cmark & &skinning+deformation  & 3DGS     \\


OmniRe~\cite{chenOmniReOmniUrbanSceneReconstruction2024a} &ICLR'25 &multi-view  &6DoF &  &\cmark &\cmark &rigid+articulated+non-rigid  &3DGS     \\

\bottomrule 
\end{tabular}
\caption{
Selected papers for dynamic scene reconstruction with hybrid motion.
}
\label{tab:hybrid}
\end{table*}

\textbf{Point Tracking} %
models each point's movement as a continuous trajectory, a time-dependent function that directly describes position at any moment in the continuous spacetime domain~\cite{liSpacetimeGaussianFeatureSplattingRealtime2024,wang2021neural}. This global trajectory formulation represents the complete motion path as a time-modulated function, eliminating accumulated errors from sequential transformations, like Eq.~\ref{cano_conn} in canonical spaces or Eq.~\ref{flow_conn} in frame-to-frame flow field. With this approach, a point's geometry and appearance can be modeled as time-varying parameters while maintaining temporal consistency~\cite{wangTrackingEverythingEverywhereAllOnce2023}.  

For multi-view capture scenarios, methods based on 3DGS can initialize the scene representation from the first frame and subsequently track each Gaussian primitive's movement through space over time~\cite{stearns2024dynamic,luiten2024dynamic}. This tracking approach can maintain consistent properties like opacity and color while updating positions, rotation, and scale, providing a more robust foundation for dynamic scene reconstruction with greater temporal coherence.

\textbf{Factorization} %
emerged from static scene reconstruction and had been successfully extended to 4D dynamic scenes through hyperplane-based factorization. Hexplane~\cite{caoHexplaneFastRepresentationDynamicScenes2023} decomposes the 4D domain into six feature planes, where point features are sampled via interpolation and concatenated to predict density and color. Similarly, K-planes~\cite{fridovich2023k} offers a unified approach for both static and dynamic scenes—factorizing static 3D space into $xy, yz, xz$ planes while representing dynamic 4D spacetime with $xt, yt, zt$ planes, incorporating multi-scale sampling for enhanced representation. This efficient factorization enables higher grid resolution and rapid convergence, making it a foundation for numerous subsequent methods~\cite{wuFastHighDynamicRangeRadiance2024a,shao2023tensor4d,linHighfidelityRealtimeNovelViewSynthesis2023,liuGearNeRFFreeViewpointRenderingTrackingMotionaware2024}. For instance, 4K4D~\cite{xu20244k4d} extends K-plane's 4D feature grid factorization to achieve real-time performance at 4K resolution, while Wu et al.~\cite{wu20244d} use factorized grid planes to encode per-Gaussian features for deformation field decoding.  

Beyond hyperplane-based factorization, basis-driven decomposition represents complex motion using a few representative spatial deformation patterns, enhancing temporal coherence while improving learning stability and storage efficiency. Li et al.~\cite{li2023dynibar} model the motion field as spatially decomposed motion bases with time-varying coefficients, using pixel-aligned features sampled from source to target views and fused by a ray transformer. While some methods represent both motion bases and coefficients as learnable neural network parameters~\cite{ramasingheBLiRFBandlimitedRadianceFieldsDynamic2024a,wangShapeMotion4DReconstructionSingle2024c}, others leverage sinusoidal bases~\cite{guoForwardFlowNovelViewSynthesis2023,wang2021neural,wangFourierPlenoctreesDynamicRadianceField2022a}.  For Gaussian-based representations, Kratimenos et al.~\cite{kratimenos2024dynmf} factorize 3D Gaussian motion into a small number of motion bases--significantly fewer than the number of Gaussian primitives--with regularization applied to the motion coefficients to ensure plausible movements. Das et al.~\cite{dasNeuralParametricGaussiansMonocularNonrigid2024} propose a two-stage approach: first learning a coarse proxy using factorized motion bases and low-rank coefficients, then initializing local volumes with 3D Gaussians refined through adaptive densification. The shared motion bases in the coarse stage force information sharing between timesteps, providing essential regularization for sparsely observed dynamic regions.

\subsection{Reconstructing with Hybrid Motion}
In autonomous driving scenes, vehicles primarily undergo rigid motion, while pedestrians and cyclists move in more complex, non-rigid ways. To address this diversity, Fischer et. al~\cite{fischer2024dynamic} leveraged tracking results with bounding boxes to represent rigid vehicles while employing non-rigid motion fields to model other dynamic objects. Taking a more comprehensive approach, OmniRe~\cite{chenOmniReOmniUrbanSceneReconstruction2024a} developed a framework that combines distinct motion representations within a dynamic scene graph: rigid nodes for vehicles, SMPL nodes for articulated pedestrian, and deformable nodes for general non-rigid objects, creating a more complete representation for urban scene reconstruction.  

Human avatar reconstruction presents another challenging hybrid motion scenario, particularly when modeling dynamic elements like hair and clothing alongside the articulated human body. Generally, there are two primary strategies for combining non-rigid deformation with articulated skinning motion. The first approach operates in canonical space, where points from observation space are initially transformed via inverse skinning, and then non-rigid deformation is applied within the canonical space~\cite{jiang2022neuman,weng2022humannerf,liu2021neural,guo2023handnerf}. The alternative approach works in observation space, first transforming the canonical body to observation space through forward skinning, then applying non-rigid displacement under the target pose~\cite{li2022tava,moon2024expressive,wen2024gomavatar}. These deformation fields typically take encoded time stamps or pose parameters as inputs to the neural networks, allowing them to capture time- and pose-dependent displacements effectively. In MonoHuman~\cite{yu2023monohuman}, a hybrid approach combines forward and inverse skinning motions with two separate non-rigid deformation fields, representing the non-rigid displacements both in target and observation space.

\subsection{Auxiliary Information and Regularization}
\label{sec.auxi}

\subsubsection{Auxiliary Information}

{\textbf{Depth Information}}
serves as a crucial geometric cue for 3D scene reconstruction, particularly valuable in dynamic scenarios where it helps mitigate ambiguities in scale, motion, and geometry. Through gradient backpropagation, supervising the rendered depth in observation space significantly aids the learning of both radiance fields and motion fields~\cite{caiNeuralSurfaceReconstructionDynamicScenes2022,wang2022learning}, particularly beneficial in challenging outdoor environments~\cite{citation-0,wu2023mars,tonderski2024neurad,yang2023unisim}. When dedicated depth sensors are unavailable, monocular depth estimation methods~\cite{ranftl2020towards,yang2024depth} can provide valuable geometric cues despite scale ambiguity~\cite{liu2023robust,wang2024gflow,liuMoDGSDynamicGaussianSplattingCausuallycaptured2024b,wang2022learning}. Beyond providing additional geometric supervision, depth information also enables importance sampling in regions near object surface, substantially reducing unnecessary computational overhead in volume-based rendering methods~\cite{deng2022depth, turki2023suds}.

\textbf{Surface Normals}
capture fine-grained geometric details that might be missed in neural representations, making them valuable for high-fidelity reconstruction. In dynamic scenes, surface normals are particularly useful in tracking shape deformations~\cite{prokudinDynamicPointFields2023,wen2024gomavatar} and modeling view-dependent reflections~\cite{yanNerfdsNeuralRadianceFieldsDynamic2023}. Recent research has successfully leveraged normal information to recover detailed human surface geometry~\cite{xiu2023econ,xiu2022icon,wang2020normalgan}, and enhance scene reconstruction~\cite{guo2024enhancing,li2024dngaussian,turkulainen2025dn}.  For supervision purpose, pseudo ground truth normals can be derived using foundation models like Metric3D~\cite{yin2023metric3d} or calculated from template meshes~\cite{liu2021neural,geng2023learning,karunratanakul2023harp}.

\textbf{Semantic Information} 
serves as a powerful auxiliary cue for accurate modeling of moving objects. In dynamic scenes, object-level semantic segmentation provides valuable silhouettes or foreground masks that help localize dynamic objects and decompose scenes into static backgrounds and dynamic foregrounds~\cite{driess2023learning,yanNerfdsNeuralRadianceFieldsDynamic2023,caiNeuralSurfaceReconstructionDynamicScenes2022}. This semantic decomposition is particularly valuable for complex urban environments with multiple moving entities~\cite{kundu2022panoptic,wu2023mars}. Beyond object-level segmentation, part-level semantic information recognizes the rigid components of articulated objects~\cite{zielonka2023drivable}, enforcing part consistency throughout the dynamic reconstruction process and enabling more detailed part-wise reconstruction~\cite{zhao2024sg}. At the finest granularity, pixel-level semantic features facilitate robust point tracking and establish reliable correspondences across frames~\cite{liang2023semantic,sinha2023common,zhou2024dynpoint,remelli2022drivable}.

\textbf{Data-driven Priors} 
significantly enhance dynamic scene reconstruction by leveraging implicit knowledge from large-scale datasets to provide valuable constraints during optimization. For example, optical flow models like UniMatch~\cite{xu2023unifying} establish dense correspondences between frames~\cite{yang2022banmo,yang2021lasr,yang2021viser,liu2022devrf}, offering important cues for flow field supervision, while object tracking models generate reliable trajectories for rigid objects in motion. Additionally, visual foundation models like DINO~\cite{oquab2024dinov2} extract semantic features that maintain consistency across both spatial viewpoints and temporal frames, providing robust cues for establishing correspondences in challenging scenarios~\cite{yang2023emernerf,guo2023handnerf,wu2023magicpony,yao2022lassie}. These data-driven approaches substantially enhance reconstruction fidelity in complex and dynamic environments where manual annotations remain prohibitively labor-intensive to acquire.

\subsubsection{Regularization}

Physical constraints play a crucial role in dynamic scene reconstruction by enforcing motion continuity and structural preservation. Temporal and spatial smoothness is achieved through Total Variation (TV) loss, which encourages piecewise constant motion with natural transitions between different regions~\cite{lombardiNeuralVolumesLearningDynamicRenderable2019,fridovich2023k,huang2024textit}, while Laplacian regularization penalizes sharp gradient changes in deformation fields~\cite{habermann2021real}. Structural integrity during deformation is maintained through As-Rigid-As-Possible (ARAP) constraints, which preserve local neighborhood relationships, and isometric loss, which maintains global geodesic distances. Additionally, divergence loss promotes volume preservation by constraining deformations to primarily consist of translations and rotations, a critical property for realistic object and scene modeling.  

Scene priors further enhance reconstruction quality by incorporating domain knowledge into the optimization process. $L_1$ regularization on motion fields biases scenes toward static components~\cite{xian2021space,wu2022d}, while cycle consistency enforces coherent correspondences between features, geometry, and appearance across frames~\cite{duNeuralRadianceFlow4dView2021,liNeuralSceneFlowFieldsSpacetime2021,yang2023emernerf}. For visual quality improvement, opacity regularization promotes binary outcomes (0 or 1) to eliminate floating artifacts~\cite{xian2021space,wu2022d}. These complementary regularizations collectively ensure that dynamic scene reconstructions achieve physical accuracy, temporal consistency, and visual plausibility.

\begin{figure}
    \centering
    \includegraphics[width=1.0\linewidth]{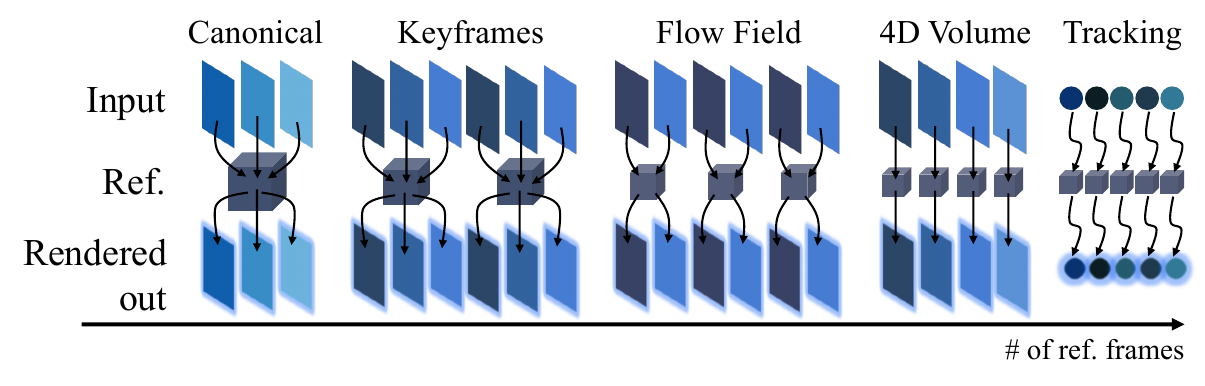}
    \caption{We propose a unified framework to encapsulate various representation paradigms.}
    \label{fig:ref_num}
\end{figure}

\subsection{Discussion}
Dynamic scene reconstruction presents significant challenges due to the complexity of capturing both spatial and temporal variations. To better analyze the diverse approaches in this domain, we propose a unified framework that encapsulates these methods by conceptualizing any dynamic scene as static reference frames with corresponding transformations to target frames. This framework provides a systematic way to categorize reconstruction techniques based on the number of reference frames employed, as illustrated in Fig~\ref{fig:ref_num}.  

For scenes exhibiting rigid or articulated motion, a single canonical space is typically sufficient to represent both geometry and appearance. Similarly, scenes with relatively simple non-rigid deformations can often be reconstructed using a shared canonical space as a common reference. However, as sequences become longer or motions more complex, relying on a single global canonical space becomes increasingly challenging. In such cases, multiple keyframes serving as local references provide a more effective solution, allowing the scene to be reconstructed across several localized spaces. When this local space framework is reduced to just two frames, it transforms into the frame-to-frame flow field reconstruction paradigm, where each reference frame is mapped to its consecutive neighbor using 3D correspondence estimation. Taking this concept further, each frame can serve as its own reference in a frame-by-frame optimization approach, effectively treating the dynamic scene as a continuous 4D spacetime volume. At the finest granularity, per-point tracking establishes reference at the point level, creating individual trajectories for scene elements.  

Generally, as the number of reference frames increases and the level of granularity becomes finer, the reconstruction captures more detailed temporal dynamics. However, this enhanced detail comes at the cost of increased computational complexity. This trade-off between fidelity and practical efficiency represents a fundamental consideration when selecting the appropriate reconstruction approach for specific applications.

\section{Challenges and Future Trends}
\label{sec:future}
\textbf{Manipulability and Editability.} %
\label{sec:edit}
While substantial progress has been achieved in manipulating and editing 2D images and 3D static scenes, extending these capabilities to 4D spatiotemporal representations remains challenging. Recent approaches have demonstrated promising results in scene-level style transfer~\cite{liu2024stylegaussian,liuDynvideoeHarnessingDynamicNerfLargescale2024} and object-level manipulations (removal, addition, repositioning)~\cite{yang2023unisim,wu2023mars}, as well as decomposing complex scenes into static and dynamic components~\cite{leeCompact3DGaussianSplattingStatic2024,roldao2024rodus}. However, fine-grained part and pixel-level editing in dynamic scenes presents significant difficulties. The critical challenge lies in establishing accurate point correspondence across time to ensure temporal consistency during edit propagation. The explicit representation afforded by 3DGS has recently enabled advances in dense tracking~\cite{luiten2024dynamic}, demonstrating potential for pixel-level manipulation; nevertheless, developing structured editing paradigms for inherently unstructured radiance fields remains an open research problem.

\textbf{Scalability.} %
\label{sec: scale}
Dynamic scene reconstruction faces three critical scalability challenges: spatial extent, temporal duration, and motion complexity. Spatially, radiance fields struggle when extended to vast environments like city-level scenes, where memory requirements grow prohibitively with scene size. While divide-and-conquer strategies have been proposed~\cite{tancik2022block,lin2024vastgaussian,xu2023grid}, these approaches often struggle with integration and consistency across boundaries, particularly for dynamic scenes. Temporally, computational demands scale linearly with sequence duration, making reconstruction of extended periods (from minutes to days) increasingly prohibitive with current architectures. This challenge is compounded by the difficulties in maintaining robust long-term tracking for motion recovery, as occlusions and dramatic changes in object appearance frequently disrupt correspondence establishment. Despite recent advances, simultaneously addressing spatial scale, temporal extent, and complex non-rigid motion remains an open research challenge requiring fundamental breakthroughs in scene representation and optimization techniques.

\textbf{Reconstrution by Generation.} %
\label{sec: future data engine}
Dynamic scene reconstruction faces a fundamental challenge: while high-quality results require comprehensive visual data, practical applications often rely on casually captured monocular footage that provides severely limited information, resulting in incomplete reconstructions~\cite{gaoMonocularDynamicViewSynthesisReality2022}. Static scene reconstruction has successfully leveraged generative approaches to synthesize invisible or occluded regions using models like Latent Diffusion Models~\cite{rombach2022high}, but extending these capabilities to 4D dynamic scenes remains problematic. Such integration demands simultaneous maintenance of spatial view consistency, temporal coherence, and plausible motion dynamics. Despite recent advances in unconditional, image-guided, and text-prompted 4D content generation~\cite{podellsdxl,lin2023magic3d,raj2023dreambooth3d}, current methods predominantly produce 2D frame sequences without underlying 3D structure~\cite{liu2024sora}, failing to provide comprehensive volumetric representations. The critical research challenge lies in effectively conditioning generative models on partial inputs to produce geometrically accurate and temporally consistent 4D volumes.

\textbf{Large Language Models.}
\label{sec: future foundation model}
Large Language Models (LLMs)~\cite{chang2024survey} offer powerful semantic priors for dynamic scene understanding~\cite{singer2023text,hong2022avatarclip}, complementing visual foundation models through their world knowledge and reasoning capabilities. Despite their potential, integrating LLMs with 4D reconstruction presents significant challenges: high-fidelity reconstruction requires pixel-precise geometry and appearance modeling, while LLMs primarily provide high-level semantic abstractions difficult to align with fine-grained visual features. Recent approaches like Language Embedded 3D Gaussians~\cite{lingAlignYourGaussiansTextto4dDynamic2024,qin2024langsplat} demonstrate promising directions by incorporating quantized semantic features into explicit scene representations, enabling language-guided editing and querying of 3D content. Future research opportunities lie in developing bidirectional interfaces between LLMs' symbolic reasoning and the spatiotemporal representations required for dynamic scenes, potentially enabling physics-aware and semantically meaningful scene reconstruction and manipulation.

\section{Conclusion and Outlook}
\label{sec:conclusion}

In this survey, we present a comprehensive overview of dynamic motion and scene representation in radiance fields, focusing on Neural Radiance Fields and 3D Gaussian Splatting. By systematically categorizing motion into rigid, articulated, non-rigid, and hybrid types, we analyze diverse approaches across the literature, highlighting their strengths and limitations while identifying critical challenges and promising research directions.  

\textbf{Outlook.} While 3D scene reconstruction has achieved remarkable success, the research frontier has shifted toward comprehensive 4D dynamic volume reconstruction. Powered by advances in neural rendering, generative models, foundation models, and LLMs, we anticipate rapid progress in simultaneously addressing two fundamental challenges: photorealistic geometry and appearance reconstruction, and consistent, physically plausible temporal motion recovering. In summary, 4D dynamic scene reconstruction presents both significant opportunities and challenges, with the ultimate goal of creating high-fidelity digital twins of real dynamic physical environments.  In this era of emerging technologies, we hope this survey serves as a valuable foundation to inspire researchers pursuing advances in this field.

\section*{Acknowledgments}

This project is partially supported by NSFC (62476075 and 62376080), the Zhejiang Provincial Natural Science Foundation Key Fund of China (LZ23F030003), and the Zhejiang Key Laboratory of Optoelectronic Intelligent Imaging and Aerospace Sensing.

\bibliographystyle{ieeetr}
\bibliography{ref}

\appendix

\subsection{\textbf{More Detailed Discussion about Capture Setting}}
\begin{figure}
    \centering
    \includegraphics[width=0.95\linewidth]{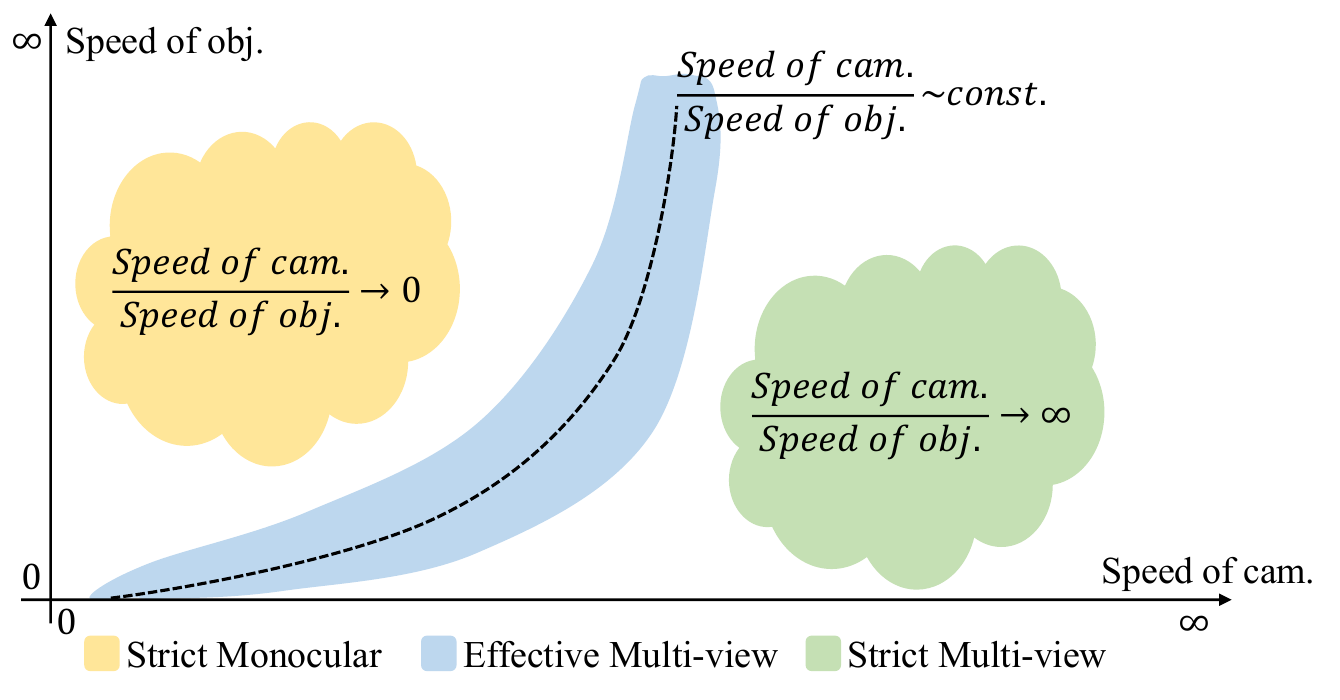}
    \caption{Relative motion speed between camera and objects determines the critical factor in effective scene capture.}
    \label{fig:capture}
\end{figure}

Dynamic scene reconstruction requires tracking points across video frames, where 2D pixel movement represents a combination of both object and camera motion. The type of sensor setup and capture strategy significantly impacts the system's ability to accurately estimate true 3D motion, ultimately determining the upper bound of reconstruction outcomes. For static scenes, a single moving camera can provide information equivalent to multiple cameras by capturing different viewpoints over time. However, when objects themselves move, the relationship between camera motion and object motion becomes crucial~\cite{gaoMonocularDynamicViewSynthesisReality2022}. This relationship creates different levels of ambiguity and reconstruction difficulty, categorized as: (1) strict monocular, where the camera moves much slower than objects, creating occlusion challenges, (2) effective multi-view, where camera and object speeds are comparable, allowing the camera to ``follow" and maintain visibility of key points, and (3) strict multi-view, where multiple synchronized cameras capture the scene simultaneously from different views. When camera movement closely matches object movement, even single-camera setups can track features effectively enough for reasonable reconstruction, as illustrated in Fig.~\ref{fig:capture}. Importantly, the critical factor is not the absolute motion speed of camera or object, but rather their relative speed ratio.

While strict multi-view camera setups theoretically provide the most complete information for high-quality reconstruction, they introduce practical challenges in synchronization, data management, and deployment complexity. Strict monocular approaches, despite capturing less complete information about the scene, offer significantly simpler solutions for real-world applications. The trade-off between capture complexity and reconstruction quality continues to shift as algorithms improve in handling limited input data, with effective multi-view approaches representing a promising middle ground, making simpler camera setups increasingly viable for many practical uses.

\subsection{\textbf{More Detailed Discussion about NeRF and 3DGS}}
Neural Radiance Fields and 3D Gaussian Splatting represent complementary approaches to volumetric scene representation, each with distinct tradeoffs. NeRF's implicit continuous representation excels at capturing fine details and complex view-dependent effects through compact network parameters, but requires computationally expensive ray marching that precludes real-time applications. In contrast, 3D Gaussian Splatting's explicit primitive-based representation enables real-time rendering through efficient rasterization, though complex scenes may demand large numbers of Gaussians that increase memory requirements. The choice between these approaches thus depends on specific application priorities—whether visual fidelity, computational efficiency, or memory constraints takes precedence.

These methods fundamentally differ in their rendering techniques: ray tracing in NeRF provides continuous sampling that accurately captures complex optical phenomena, while splatting in Gaussian representations serves as an approximation of physically-based rendering, sacrificing some fidelity in representing sophisticated light transport effects such as reflections and shadows. This complementarity has motivated recent exploration of hybrid techniques that introduce ray tracing capabilities into Gaussian-based representations to better capture secondary lighting effects while maintaining computational efficiency~\cite{byrski2025raysplats,moenne20243d}. As the field advances, developing representations that simultaneously achieve efficiency, fidelity, and real-time performance remains an active frontier in neural rendering, with each approach continuing to inform and enhance the other.

\subsection{\textbf{More Detailed Discussion about Representation Paradigm}}
Dynamic scene reconstruction presents significant challenges due to its inherently ambiguous nature, particularly when limited observations are available in monocular settings. These challenges have led to the development of various representation paradigms, each offering different trade-offs between accuracy, efficiency, and flexibility. For rigid objects, the key challenge lies in accurately aligning tracking results from different time stamps into a unified local coordinate system, enabling effective temporal information aggregation. The precision of object localization and tracking therefore becomes fundamental to successful rigid motion reconstruction. While articulated objects consist of rigidly moving parts, their overall motion is non-rigid, making part-level representation more appropriate than object-level approaches. To unify these parts coherently, category-level hierarchical structures are often introduced as priors with specific kinematic constraints. However, developing such kinematic templates is labor-intensive, resulting in templates for only a limited number of articulated motion categories (e.g., humans, quadruped animals). Consequently, template-free reconstruction of articulated motion remains a meaningful yet challenging research direction, with existing approaches showing promise but requiring further quality improvements.   

The canonical space approach represents a fundamental design choice where appearance and geometry remain static in a universal reference frame, while neural deformation fields map points between this canonical frame and observed frames. This approach enables effective scene editing with changes propagating through the deformation field, but struggles with sequences exhibiting extreme deformations or topological changes, where maintaining a single coherent canonical space becomes increasingly difficult. Multiple keyframe approaches offer a middle ground by establishing several reference frames rather than relying on a single canonical space. This paradigm relaxes the constraint of finding one universal reference while maintaining temporal coherence across subsets of frames. The extreme case is frame-by-frame optimization in 4D spacetime representation, where each frame functions as its own keyframe. While this approach can achieve high-quality individual frame reconstructions by focusing on per-frame accuracy, it fails to disentangle motion from scene representation, significantly limiting subsequent manipulation capabilities and producing temporally inconsistent results.  

Frame-to-frame flow fields bridge adjacent frames through small point displacements, effectively decomposing complex global deformations into more manageable local transformations. This approach eliminates the need for a shared canonical space and handles large deformations by breaking them into incremental steps. However, these local correspondences often lack temporal consistency over long-term sequences as small errors accumulate over time. Point tracking via trajectory fields addresses this limitation by modeling each point's path as a continuous function of time rather than discrete connections, constraining motion across an object's entire lifespan and maintaining temporal coherence throughout the video sequence. However, unrestricted trajectory functions may still yield physically implausible motions without proper regularization.

Motion factorization methods decompose complex motion into a limited set of basis trajectories with time-dependent coefficients, effectively capturing shared motion patterns while reducing the solution search space. This transforms the challenging problem of regularizing implicit motion fields into the more intuitive task of regularizing motion coefficients, often leading to more reasonable and physically plausible recovery. In practical applications, hybrid approaches combining multiple representation paradigms yield superior results, particularly for scenes with mixed motion types including rigid objects, articulated entities, and general non-rigid deformations. The learning paradigm also significantly impacts reconstruction quality, with progressive learning strategies—starting from coarse to fine details, incrementally increasing frequency bands, or employing two-stage reconstruction that separates static and dynamic elements—proving effective in handling complex dynamic scenes while avoiding local minima during optimization.




\end{document}